\documentclass[final,3p, 10pt]{elsarticle}

\usepackage{graphics}
\usepackage{graphicx}
\usepackage{amsfonts}
\usepackage{amssymb}
\usepackage{amsmath}
\usepackage{dsfont}
\usepackage{color}
\usepackage{hyperref}
\usepackage{subcaption}
\usepackage{epsfig}
\usepackage{float}

\usepackage{color}

\usepackage[ruled]{algorithm2e}
\usepackage{algorithmic}

\newcommand{\Tref}{T_{\textnormal{ref}}}
\newcommand{\Yexit}{Y_{\textnormal{e}}}
\newcommand{\Yin}{Y_{\textnormal{ref}}}
\newcommand{\Yideal}{Y_{\textnormal{e,ideal}}}

\newcommand{\bu}{\mathbf{u}}

\newcommand{\bc}{\mathbf{c}}
\newcommand{\bw}{\mathbf{w}}

\newcommand{\bI}{\mathbf{I}}
\newcommand{\bA}{\mathbf{A}}
\newcommand{\bB}{\mathbf{B}}

\newcommand{\bV}{\mathbf{V}}
\newcommand{\bU}{\mathbf{U}}

\newcommand{\bW}{\mathbf{W}}
\newcommand{\bJ}{\mathbf{J}}

\newcommand{\bxi}{\boldsymbol{\xi}}

\newcommand{\btheta}{\boldsymbol{\theta}}

\newcommand{\balpha}{\boldsymbol{\alpha}}
\newcommand{\bbeta}{\boldsymbol{\beta}}
\newcommand{\bgamma}{\boldsymbol{\gamma}}
\newcommand{\bfeta}{\boldsymbol{\eta}}

\newcommand{\bzeta}{\boldsymbol{\zeta}}

\newcommand{\calG}{\mathcal{G}}
\newcommand{\calL}{\mathcal{L}}
\newcommand{\calD}{\mathcal{D}}

\newcommand{\calP}{\mathcal{P}}

\newcommand{\calF}{\mathcal{F}}
\newcommand{\calM}{\mathcal{M}}

\newcommand{\calJ}{\mathcal{J}}

\newcommand{\R}{\mathbb{R}}
\newcommand{\N}{\mathbb{N}}
\newcommand{\E}{\mathbb{E}}
\newcommand{\Prob}{\mathbb{P}}

\newcommand\aseq{\mathrel{\stackrel{\makebox[0pt]{\mbox{\normalfont\tiny a.s.}}}{=}}}

\begin{document}

\begin{frontmatter}

\title{Compressive sensing adaptation for polynomial chaos expansions}
\date{}

\author[label0,label1]{Panagiotis Tsilifis\corref{cor1}}
\ead{panagiotis.tsilifis@epfl.ch}

\author[label2,label3]{Xun Huan}
\ead{xhuan@umich.edu}

\author[label3]{Cosmin Safta}
\ead{csafta@sandia.gov}

\author[label3]{Khachik Sargsyan}
\ead{ksargsy@sandia.gov}

\author[label3]{Guilhem Lacaze}
\ead{guilhem.lacaze@gmail.com}

\author[label3]{Joseph~C.~Oefelein}
\ead{joseph.oefelein@aerospace.gatech.edu}

\author[label3]{Habib N. Najm}
\ead{hnnajm@sandia.gov}

\author[label1]{Roger G. Ghanem} 
\ead{ghanem@usc.edu}

\cortext[cor1]{Corresponding author}
\address[label0]{CSQI, Institute of Mathematics, \'{E}cole Polytechnique F\'{e}d\'{e}rale de Lausanne, Lausanne, CH-1015, Switzerland}
\address[label1]{Sonny Astani Department of Civil Engineering, University of Southern
California, Los Angeles, CA 90089, USA} 
\address[label2]{Department of Mechanical Engineering, University of Michigan, Ann Arbor, MI 48109, USA}
\address[label3]{Sandia National Laboratories, 7011 East Avenue, Livermore, CA 94550, USA}


\begin{abstract}
Basis adaptation in Homogeneous Chaos spaces rely on a suitable rotation of the underlying Gaussian germ.  Several rotations have been proposed in the literature resulting in adaptations with different convergence properties.  In this paper we present a new adaptation mechanism that builds on compressive sensing algorithms, resulting in a reduced polynomial chaos approximation with optimal sparsity.
The developed adaptation algorithm consists of a two-step optimization procedure that computes the optimal coefficients and the input projection matrix of a low dimensional chaos expansion with respect to an optimally rotated basis. We demonstrate the attractive features of our algorithm through several numerical examples including the application on Large-Eddy Simulation (LES) calculations of turbulent combustion in a HIFiRE scramjet engine.
\end{abstract}

\end{frontmatter}

\section{Introduction}

While the use of computer codes to represent complex physical phenomena has always been an integral part of uncertainty quantification (UQ),
efforts toward adopting more realistic models often face
computational challenges that must be overcome before
meaningful quantitative analysis becomes possible.  The need for
a statistical exploration of the solution space is arguably the most
pressing of these challenges, requiring repetitive simulation of the
numerical code.  The associated computational
burden quickly becomes prohibitive, especially in the context of
complex physical systems where each of these simulations is, by
itself, already testing the limits of computational resources. In many
important instances, the map from input random parameters to output
quantities is highly nonlinear, limiting the value of standard
statistical sampling techniques. In such
cases, $L_2$-based formalisms, such as polynomial chaos (PC) expansions
have shown promise \cite{ghanem_spanos,le_maitre_etal,reagan,najm}, though, by explicitly tracking each input stochastic
parameter, they are subject to the curse of dimensionality \cite{bellman}. The latter is
manifested by a factorial growth of the number of parameters and therefore result in an overwhelming increase of the numerical simulations required to systematically explore the parameter space.
Significant efforts have been
expanded recently to leverage the mathematical structure provided by
an $L_2$ resolution to alleviate that computational burden with
rigorous and tractable error control  \cite{Doostan:2007,Arnst:2012,tipireddy}.
While no single approach works universally, successful model reduction
strategies have addressed these challenges with either the
construction of a model surrogate that replaces the
expensive initial model \cite{ghanem_spanos,rasmussen}, or the representation of the model output as
a mapping from a reduced input space \cite{saltelli, pearson, Doostan:2007, Arnst:2012,tipireddy,constantine, tripathy}.

In this paper, we concentrate our efforts on approximating \textit{nonlinear} response surfaces by polynomial chaos expansions that consist of \textit{linear} series of terms that are orthogonal with respect to the
probability measure of input
variables.  The foundation of these representations was pioneered in
the context of stochastic finite elements \cite{ghanem_wrr, ghanem,
  ghanem_dham, ghanem_redhorse, ghanem_spanos, xiu_fluid, najm}, and the
Hilbertian structure of the underlying $L_2$ space permitted the
development of non-intrusive approaches for estimating the expansion
coefficients \cite{le_maitre_etal, babuska, marzouk_etal}. More
recently, basis enrichment \cite{Ghosh:2008}, least-squares
\cite{blatman, doostan_icc, tsilifis_design} and $\ell_1$-minimization methods
\cite{Candes2006a, donoho, doostan_owhadi, peng_l1} were
suggested as enhancements or alternatives to the spectral perspective.
Further, recent advances of
$\ell_1$-minimization have demonstrated additional sparsity enhancement
by adaptively selecting the basis terms \cite{sargsyan_IJUQ,jakeman}.

A recently introduced dimension reduction technique consists of a basis
adaptation procedure \cite{tipireddy} which constructs polynomial
chaos expansions for specific quantities of interest (QoIs)
using only a small number of Gaussian variables
that are linear combinations of the original basis.  These combinations are
specifically adapted to the QoI in question and are obtained through a
learning process which involves exploring the solution space through a
limited number of samples.
Standard basis adaptation procedures involve choosing these linear
combinations through a rotation matrix computed according to
particular sparse grid rules, following a low-dimensional quadrature
rule for evaluating the adapted expansion for the QoI. While basis
adaptation was shown to be effective for reduced representation of random fields
\cite{tsilifis} and design optimization under uncertainty
\cite{thimmisetty}, the overall computational cost (while linear in
the dimension) can be further reduced.

The main contribution of the paper is a novel algorithm that
efficiently and simultaneously computes the basis rotation as well as
the corresponding chaos coefficients
using a fixed number of model evaluations independent of the choice of
reduced dimensionality, and as a result departing from the restrictive
traditional pseudo-spectral approaches.  This is achieved by
incorporating an $\ell_1$-minimization procedure on Hermite Chaos
expansions with respect to variables that are assumed to be orthogonal
projections of the original input variables through a projection map
that is computed jointly with the $\ell_1$-minimization. As was emphasized in the original basis adaptation method \cite{tipireddy}, our approach applies specifically to the Hermite Chaos with Gaussian input variables, as the distribution of the projected variables would otherwise be arbitrary, resulting in non optimal polynomial representation, due to the loss of the orthogonality property. A recent attempt to adapt the basis of non-Hermite Chaos \cite{tsilifis_gradient}, has been restricted to projections on 1-dimensional bases, as those can be easily mapped to uniformly distributed inputs and the Legendre Chaos can then be employed. This however required an \emph{a priori} indication that such a 1-dimensional adaptation exists, which was validated using a gradient-based criterion to compute the rotation, that again relies on pseudo-spectral approaches and we therefore restrain from such exploitations here. The
advantages of our algorithm are highlighted on an example of extreme
scale computation for a realistic engineering application, involving
large-eddy simulations (LES) of supersonic turbulent reactive flows
inside a scramjet engine combustor, where the input space is high
dimensional and a very limited number of these expensive simulations
is available.

Few previous works for achieving dimensionality reduction within the context of PC or generic polynomial surrogates are known to the authors. Namely, a similar heuristic algorithm has been proposed in the past \cite{yang_sparsity}, where the combination of an $\ell_1$-minimization approach for computing the chaos coefficients together with the active subspace method for estimating a rotation matrix, has resulted in improved sparsity in the PC expansion. Furthermore, an approach for generic polynomial ridge functions was proposed in \cite{hokanson} where both the coefficients and the projection matrix are estimated using least squares minimization. First, for the coefficients, their well-known least squares solution that involves a Moore-Penrose pseudoinverse of the measurement matrix, was substituted in the objective function. The implicit dependence of the pseudoinverse on the projection matrix results in replacing the two step procedure by a single optimization problem that is referred to as \emph{variable projection} approach. Our work, as is explained below, offers a new alternative that retains the benefits of a sparse solution ensured by the use of $\ell_1$-minimization as in the first reference, while the least squares solution for the rotation matrix allows for a data-driven adaptation as in the second approach.

This paper is structured as follows. Section \ref{sec:poly_chaos}
describes the use of PC expansion as a response surface, specifically the
Hermite (Homogeneous) Chaos for both standard Gaussian variables and
rotated Gaussian variables produced from the basis adaptation
procedure.
Section \ref{sec:comp_sens} provides the main ingredients of
compressive sensing, which are combined with basis adaptation to
estimate the new expansion coefficients.
The overall method is then demonstrated on a series of numerical
examples in Section \ref{sec:examples}, including a 12-dimensional
ridge function,
a 20-dimensional Burgers' equation,
and an 11-dimensional scramjet combustor application.
The paper then ends with conclusions in Section \ref{sec:conclusions}.

\section{Methodology}

\subsection{Polynomial Chaos Expansion}
\label{sec:poly_chaos}

\subsubsection{Homogeneous Chaos}
\label{sec:homo_chaos}

Throughout this paper, let us assume the quantity
\begin{equation}
u := u(\bxi),
\end{equation}
that can be written as a function of uncorrelated Gaussian variables $\bxi = (\xi_1, \dots, \xi_d)$, is a square integrable function, that is
$u \in L^2(\Omega, \calF, \Prob)$, where $\calF := \calF(\calG)$ is the $\sigma$-algebra generated from the Gaussian Hilbert space $\calG = \mbox{span}\{\xi_i\}_{i=1}^d$. It is known \cite{wiener, cameron, ghanem_spanos} that $u$ admits a
series expansion of the form
\begin{equation}
\label{eq:chaos_expansion}
u(\bxi) = \sum_{\balpha, |\balpha| = 0}^{\infty} c_{\balpha} \psi_{\balpha}(\bxi),
\end{equation}
where $\balpha = (\alpha_1, \dots, \alpha_d) \in \calJ^d := \N^d \cup
\{\mathbf{0}\}$ are finite-dimensional multiindices with norm
$|\balpha| = \alpha_1 + \dots + \alpha_d$, and the basis functions $\psi_{\balpha}$ are defined as the tensor product
\begin{equation}
\psi_{\balpha}(\bxi) = \prod_{i = 1}^d \psi_{\alpha_i}(\xi_i)
\end{equation}
with
\begin{equation}
\label{eq:norm_hermite}
\psi_{n}(\xi) = \frac{h_n(\xi)}{\sqrt{n!}} \end{equation}
and $h_n$ is the standard $1$-dimensional Hermite polynomials of order $n$ which is orthogonal with respect to the Gaussian measure with density $p(\xi) = (2\pi)^{-1/2}\exp(- \xi^2/2)$ and has norm $\Vert h_{n}\Vert = 1\ , \ n \in \N$. The Hilbert structure of $L^2(\Omega, \calF, \Prob)$ is characterized by the inner product defined as 
\begin{equation}
\langle \psi_{\balpha}(\bxi), \psi_{\bbeta}(\bxi) \rangle := \E\left\{\psi_{\balpha}\psi_{\bbeta}\right\} = \int_{\R^d}
\psi_{\balpha}(\bxi)\psi_{\bbeta}(\bxi) p(\bxi) d\bxi 
\end{equation}
where $p(\bxi) = \prod_{i=1}^d p(\xi_i)$, $||\psi_{\balpha}|| =
\left(\E\{\psi_{\balpha}^2\}\right)^{1/2}$, thus the orthogonality condition is given by 
\begin{equation}
\langle \psi_{\balpha}(\bxi), \psi_{\bbeta}(\bxi) \rangle =
||\psi_{\balpha}||^2 \delta_{\balpha, \bbeta},
\end{equation}
where $\delta_{\balpha,\bbeta}$ is the Dirac delta function taking the value
of $1$ if $\balpha = \bbeta$ and $0$
otherwise. Eq. (\ref{eq:norm_hermite}) suggests that
$||\psi_{\balpha}|| = 1$, and so the polynomials are normalized. We
refer to Eq. (\ref{eq:chaos_expansion}) as the \emph{polynomial chaos
  expansion} of $u$.

In practice, we work with truncated versions of
(\ref{eq:chaos_expansion}). For $Q \in \N$, $\calJ^d_Q :=
\left\{\balpha \in \calJ^d:\ |\balpha|\leq Q\right\}$, we assume that
$u$ can be \emph{accurately} approximated by
\begin{equation}
\label{eq:chaos_trunc}
u(\bxi) \approx \sum_{\balpha\in \calJ^d_Q} c_{\balpha}\psi_{\balpha}(\bxi).
\end{equation}
This truncated expansion of order $Q$ consists of
\begin{equation}
N_Q = \left(\begin{array}{c}d + Q \\ Q\end{array}\right) = \frac{(d + Q)!}{d! Q!}
\end{equation}
basis terms whose coefficients $\{c_{\balpha}\}_{\balpha\in \calJ^d_Q}$ need to be computed.

\subsubsection{Adaptation on the Gaussian basis}
\label{sec:basis_adapt}

From the above, it is clear that all $u$ that are $\calF$-measurable
can be expressed as a function of any basis of $\calG$. This
encompasses any set of uncorrelated standard normal random variables
that spans $\calG$, since the latter generates identical Chaos spaces
of higher order. Assume $\bA : \R^d \to \R^d$ is a unitary matrix
($\bA \bA^T = \bI$) that serves as a linear operator from $\R^d$ to
itself, and taking $\bxi$ to be an \emph{initially} chosen basis, then
\begin{equation}
\bfeta = \bA \bxi
\end{equation}
defines a new set of independent standard normal random variables that
spans $\calG$, and therefore generating the same $\sigma$-algebra
$\calF(\calG)$. As a result, any $u\in L^2(\Omega, \calF, \Prob)$ can
also be expanded as
\begin{equation}
\label{eq:chaos_BA}
u := u(\bfeta) = \sum_{\bbeta \in \calJ_Q^d} \tilde{c}_{\bbeta} \psi_{\bbeta}(\bfeta) = \sum_{\bbeta \in \calJ_Q^d} \tilde{c}_{\bbeta} \psi_{\bbeta}(\bA\bxi)
\end{equation}
where from the almost sure equality $u(\bxi) \aseq u(\bfeta)$ we have that 
\begin{equation}
c_{\balpha} = \sum_{\bbeta \in \calJ_Q^d} \tilde{c}_{\bbeta} \langle\psi_{\bbeta}(\bA\bxi), \psi_{\balpha}(\bxi) \rangle.
\end{equation}

Of high interest is the $\bA$ that leads to an expansion of
$u(\bfeta)$ (for a given fixed order $Q$) that depends primarily only
on a small number of components. In other words, we would like to
construct $\tilde{\bfeta} = (\eta_1, \dots, \eta_{d_0})^T$ with $d_0
\ll d$ such that
\begin{equation}
\label{eq:chaos_ba_W}
u(\bxi) \approx u(\tilde{\bfeta}) = u(\bW \bxi) = \sum_{\bgamma \in \calJ_Q^{d_0}} \tilde{c}_{\bgamma} \psi_{\bgamma}(\bW \bxi),
\end{equation}
where the coefficients of the terms $\psi_{\bgamma}(\bfeta)$ for
$\bgamma \in \calJ_Q^d \smallsetminus \calJ_Q^{d_0}$ are assumed to
take small values and therefore can be neglected. Here, $\bW$ is the
matrix from decomposing the isometry
\begin{equation}
\bA = \left[\begin{array}{c}
\bW \\ \bV
\end{array}\right]
\end{equation}
where $\bW^T \in \calM_{d_0}^d$, $\bV^T \in \calM_{d-d_0}^d$ with $\calM_{m}^n$ being the set of $n\times m$ matrices with orthogonal columns
\begin{equation}
\calM_{m}^n = \{\bU \in \R^{n\times m}: \bU^T \bU = \bI_m\},
\end{equation}
and is also known as the \emph{Stiefel manifold} \cite{wen}.

Several criteria for choosing the isometry $\bA$ have been proposed in
\cite{tipireddy}, but relying on knowing either the QoI cumulative
distribution function or its low (e.g., first or second) order PC
coefficients in a $\bxi$-expansion. Both approaches require prior
computations to construct $\bA$, which do not provide information on
the reduced dimensionality $d_0$, and can be computationally
inefficient
 as they are mainly associated with
 non-intrusive pseudo-spectral methods.
Our goal is to develop a novel
way of simultaneously computing optimal projection matrices and
estimating the resulting expansion coefficients, with the flexibility
of utilizing non-structured samples instead of quadrature
nodes.

\subsection{Compressive Sensing}
\label{sec:comp_sens}

\subsubsection{$\ell_1$-minimization for polynomial regression}
\label{sec:l1_minim}

To estimate the chaos coefficients $\bc = \{c\}_{\alpha \in \calJ_Q}$,
we employ compressive sensing (CS) techniques \cite{Candes2006a,
  donoho} that seek sparse PC representations. CS is particularly
  advantageous for scenarios where $\bc$ is indeed sparse, $\bxi$ is
  high-dimensional, and a very
limited number of model evaluations are available.  These methods make
use of the fact that a PC expansion is linear with respect to its
coefficients:
\begin{equation}
\bu \approx \Psi \bc,
\end{equation}
where $\bu = (\hat{u}^{(1)}, \dots, \hat{u}^{(N)})^T$ is the vector of
output data, $\{\bxi^{(i)}\}_{i=1}^N$ is the set of input points
corresponding to the data outputs, and $\Psi$ is the measurement
matrix with entries $(\Psi)_{ij} = \psi_{j}(\bxi^{(i)})$, $i =
1,\dots, N$, $j\in \calJ_Q^d$. We also denote the full dataset with $\calD =
\left\{\{\bxi^{(i)}\}_{i=1}^N ,
\{\hat{u}^{(i)}\}_{i=1}^N\right\}$, that is the set of all available data points. In practice, as we will see next, the training data, that is the data points used to infer any parameters of interest, will be either $\calD$ or a subset of it. The influence of the coefficients
is typically observed to quickly decay with higher order polynomials,
an effect that makes the $\ell_1$-minimization a suitable method when
one is interested in obtaining a sparse solution.

We focus on the following form of $\ell_1$-minimization:
\begin{eqnarray}
\label{eq:l1_problem}
\calP_{1,\epsilon}:= \left\{\arg\min_{\bc} ||\bc||_1 \ \ \ \ s.t. \ \ \  ||\bu - \Psi \bc||_2 \leq \epsilon \right\}.
\end{eqnarray}
The $\calP_{1, \epsilon}$ problem is known as the Basis Pursuit
Denoising problem. When $\epsilon = 0$ is chosen to enforce an exact
fit on the data, it is known as Basis Pursuit problem. Equivalence
with the Least Absolute Shrinkage Operator (LASSO) \cite{tibshirani}
problem can also be shown under proper choices of the regularization
and tolerance parameters \cite{donoho_elad}.

\subsubsection{Cross validation for choosing $\epsilon$}
\label{sec:cross_v}

In order to obtain a solution for the $\calP_{1, \epsilon}$ problem
that is useful for subsequent predictions, one needs to choose
$\epsilon > 0$ properly to avoid overfitting or underfitting
the data. Small values of $\epsilon$ might result in overfitting the
training data without necessarily providing accurate predictions on
points outside the training set. Large values of $\epsilon$ on the
other hand will penalize heavily on the sparsity of the solution
without taking into account the observations. We use cross-validation
to find a suitable choice of $\epsilon$. We divide the $N$ observations
into two sets consisting of $N_{tr}$ and $N_v$ samples ($N = N_{tr}+N_v$)
that will serve as the training and validation data respectively and
we denote with $\Psi_{tr}$ and $\Psi_v$ the corresponding measurement
matrices. We solve $\calP_{1,\epsilon}$ using only the $N_{tr}$
training data points and for a discrete set of values $\epsilon_{tr}$ to
obtain $\bc_{tr}$. For each solution we compute the validation error
$\epsilon_v = ||u_v - \Psi_v\bc_{tr}||_2$ and choose $\epsilon_{tr}$
such that $\epsilon_v$ is minimized. The procedure is summarized in
Algorithm \ref{alg:cv}. Alternative cross validation procedures can be preferable when large datasets are considered. These particularly involve partitioning the data into $K$ sets (folds), each consisting of $n_k = N / K$ points and repeat the above procedure $K$ times, where each time one fold serves as the validation set while the remaining points are the training set (leave-$n_k$-out cross validation) \cite{jakeman, huan_CS}. Such procedures, however, are beyond our scope.

\begin{algorithm}[h]
\caption{Cross validation algorithm for estimation of $\epsilon$ \label{alg:cv}}
Arbitrarily choose $N_{tr}$ out of $N$ data points in $\calD$, denote it with $\calD_{tr}$ and set $\calD_v = \calD \smallsetminus \calD_{tr}$. Choose a span of $J$ values $\{\epsilon_{tr}^j\}$, $j=1,\dots,J$ \\
\For{j = 1 to J}{
 $\bc_{tr} \longleftarrow$ Solution of $\calP_{1, \epsilon_{tr}^j}$ using data $\calD_{tr}$ \\
 Compute $\epsilon_v^j = ||u_v - \Psi_v\bc_{tr}||_2$.
 }
Return $\epsilon = \sqrt{\frac{N}{N_{tr}}}\epsilon^*$ where $\epsilon^* = \min_j \epsilon_v^j$. \end{algorithm}

In the above algorithm note that the $\sqrt{\frac{N}{N_{tr}}}$ scaling is
motivated by the fact that \emph{the validation error on the validation samples
becomes large as the values of $\epsilon_{tr}$ increase, while it is smaller
than the error $||\bu - \Psi \bc_{tr}||_2$ when using the full set $\calD$ \cite{doostan_owhadi}}. 
\subsubsection{$\ell_1$-minimization using adapted PCE}
\label{sec:comp_sens_BA}

Assuming now that the observed model output admits a representation of the form (\ref{eq:chaos_ba_W}), one might be interested in finding the \emph{best} projection matrix $\bW$ such that the observed data can be explained as emerging from a $d_0$-dimensional PC expansion over polynomials of $\tilde{\bfeta}$, for a given $d_0 \ll d$. The linear model in this case is written as
\begin{equation}
\bu \approx \Psi_{\bW}\bc.
\end{equation}
Here $\bu$ and $\bc$ are as in (\ref{eq:l1_problem}) while the measurement matrix has entries $\left(\Psi_{\bW}\right)_{ij} = \psi_j(\tilde{\bfeta}^{(i)})$, where $\tilde{\bfeta}^{(i)} = \bW \bxi^{(i)}$, $i = 1,\dots, N$, $j\in \calJ_Q^{d_0}$. The $\ell_1$-minimization problem can  be restated as
\begin{equation}
\label{eq:l1_ba}
\calP_{1,\epsilon}^{\bW}:= \left\{\arg\min_{\bc} ||\bc||_1 \ \ \ \ s.t. \ \ \  ||\bu - \Psi_{\bW} \bc||_2 \leq \epsilon \right\}
\end{equation}
where with $\calP_{1,\epsilon}^{\bW}$ we emphasize the dependence of the solution on the projection matrix $\bW$.

In practice, the projection matrix is not known a priori and needs to be estimated using a criterion that will guarantee some sense of optimality. Provided that all we have available is the data set $\calD$, a natural choice is to minimize the $\ell_2$ error of the model fit to the data, that is to solve \begin{equation}
\label{eq:}
\bW^* = {\arg\min}_{\bW:\bW^T \in \calM_{d_0}^d} || \Psi_\bW\bc - \bu ||_2,
\end{equation}
where $\bW$ appears only in the measurement matrix $\Psi_{\bW}$ and 
we assume that a candidate for $\bc$ (e.g., an initial guess) is available. This
motivates an iterative procedure, to be described in the next section. To further
justify our choice, it can be easily shown that this criterion is
equivalent to the maximum likelihood estimate in the Bayesian context
\cite{sargsyan_IJUQ}, see \ref{sec:map} for details. We emphasize that the above is a constrained
optimization problem since the unknown parameters are required to
satisfy the orthonormality conditions; in other words, the solution is
restricted within the Stiefel manifold $\calM_{d_0}^d$.

\subsubsection{Computational algorithm}
\label{sec:algo}

We have described the $\ell_1$-minimization problem for adapted PC
expansions that requires knowledge of $\bW$, while the estimation of
$\bW$ requires the knowledge of $\bc$. In what follows, we propose a
two-step optimization scheme that can address the challenge of solving
this coupled optimization problem. The algorithm is simply based on
the idea that the two optimization problems can be
interchangeably solved such that the solution of the one is kept fixed
while solving the other, until some convergence criterion is
satisfied. Although, at first sight, this two step approach appears to be quite heuristic, it can, in fact, be interpret as a coordinate descent algorithm that converges to the maximum a posteriori solution corresponding to a Bayesian formalism of the problem, see \ref{sec:map} for detailed explanation.
The pseudocode for this idea is summarized in Algorithm
\ref{alg:CS_BA}.

\begin{algorithm}[h]
\caption{Compressive sensing with built-in basis adaptation \label{alg:CS_BA}}
\SetKwInOut{Input}{Input}\SetKwInOut{Require}{Require}
\DontPrintSemicolon
\Require{Observed inputs $\{\bxi^{(i)}\}_{i=1}^N$, observed outputs $\{u^{(i)}\}_{i=1}^N$, choice of $d_0 < d$, initial guess $\bc^0 \in \R^{|\calJ_Q^{d_0}|}$, $\bW^0 \in \calM_{d_0}^d$, maximum number of iterations $M_{iter}$, convergence tolerances $\epsilon_{\ell_1}$ and $\epsilon_l$, fitting error $\epsilon$. Set $it = 1$.}
\Repeat{ relative change in $||\bc||_1$ is less than $\epsilon_{\ell_1}$ and change in $\calF(\bW)$ is less than $\epsilon_l$ or $it = M_{iter}$}{
        Compute $\{\bfeta^{(i)}\}_{i=1}^N$ and $\Psi_{\bW^{it-1}}$ where $\bfeta^{(i)} = \bW^{it-1} \bxi^{(i)}$, $i = 1, \dots, N$  and $\left[\Psi_{\bW^{it-1}}\right]_{kl} = \psi_{l}(\bfeta^{(k)})$\\
        $\bc^{it} \leftarrow \arg\min_{\bc}||\bc||_1$ subject to $||\bu - \Psi_{\bW^{it-1}} \bc||_2 < \epsilon$\\
        $\bW^{it} \leftarrow \arg\min_{\bW} ||\bu - \Psi_{\bW} \bc^{it}||_2^2$\\
        $it \leftarrow it + 1$
}
\end{algorithm}

While the proposed algorithm involves iterating between two tractable
subproblems ($\ell_1$ and $\ell_2$-minimizations), it does not address
how to choose $d_0$, and the issue of increasing dimensionality of
both arguments $\bc$ and $\bW$ when one increases $d_0$. More
specifically, upon solving (\ref{eq:l1_ba}) for a small value $d_0$,
one may decide that the resulting PC expansion is not accurate enough,
and therefore, the need to increase $d_0$ and repeat the
procedure. The number of expansion coefficients increases
factorially with $d_0$ while the number of entries in $\bW$
increases geometrically, and the combined effect can result in an
expensive-to-solve problem as we move to larger $d_0$ values. In
practice, the growth mainly affects the constrained optimization
problem with respect to $\bW$, and the convergence to a global minimum
can become slow.

Another drawback of the proposed procedure is the possibility to be
stuck in a local minimum. This is mainly due to the fact that the objective function to be optimized with respect to $\bW$ is generally non-convex. 
In addition, it can be observed that for a
given $d_0$, the optimal solution provides a PC expansion with respect
to a germ $\bfeta$ that can itself be rotated along the
$d_0$-dimensional space, resulting in an infinite number of possible
expansions that are almost surely equal and with the same $\ell_2$
value, while the $\ell_1$ norms of the chaos coefficients are not
necessarily equal. This property is further explained in
\ref{sec:equiv}. As a result, the algorithm might not converge to the
global maximum likelihood value when minimizing with respect to $\bW$.

In order to reduce the number of parameters in our optimization problem, and thus improve its efficiency, we also propose a second algorithm that
computes the rows of $\bW$ by successively solving the optimization
problem with respect to each row at a time while fixing the entries of
the rows that have already been estimated; the pseudocode is
presented in Algorithm \ref{alg:CS_BA_2}. This algorithm replaces the
problem of minimizing the $\ell_2$ error with respect to $d_0\times d$
parameters with that of solving $d_0$ minimization problems with $d$
parameters each time (note that the increase of the number of chaos coefficients at
each problem does not add up significant computational complexity). While this new variant mitigates the computational burden at each iteration, the challenge of local minima remains.
To further assist convergence to the global minimum, we repeat the
procedure multiple times from different initial conditions and select
the solution corresponding to the lowest minimum.
At small values of $d'$, it is possible that the linear system is
overdetermined and ordinary least squares (OLS) can be employed instead of
$\calP_{1, \epsilon}$. One can therefore replace the corresponding
step with an OLS solution until the problem becomes underdetermined as
$d'$ is increased. Both approaches are expected to perform similarly
as the $\calP_{1,\epsilon}$ solution tends toward the OLS solution
for large values of $N$.
\begin{algorithm}[h]
\caption{Successive row estimation of the projection matrix \label{alg:CS_BA_2}}
\SetKwInOut{Initialize}{Initialize}\SetKwInOut{Require}{Require}
\DontPrintSemicolon
\Require{Observed inputs $\{\bxi^{(i)}\}_{i=1}^N$, observed outputs $\{u^{(i)}\}_{i=1}^N$, choice of $1 < d_0 < d$, initial guess $\bc^0 \in \R^{|\calJ_Q^{1}|}$, $\bw^0 \in \calM_{1}^d$, maximum number of iterations $M_{iter}$, convergence tolerances $\epsilon_{\ell_1}$ and $\epsilon_l$, fitting error $\epsilon$.}
\Initialize{Set $d' = 1$ and use Algorithm \ref{alg:CS_BA} to solve (\ref{eq:l1_ba}) and obtain $\bW^* \in \calM_{1}^d$ and $\bc^* \in \R^{|\calJ_Q^{1}|}$.}
\For{ $d' = 2$ to $d_0$}{
        $\cdot$ Set $\bW^{d'} = \bW^*$.\\
        $\cdot$ Generate random initial guess $\bc^0 \in \R^{|\calJ_Q^{d'}|}$, $\bw^0 \in \calM_{1}^d$ such that $\bW^{d'} \cdot \bw^{0T} = \mathbf{0}\in \R^{d'-1}$ and set $\bW^0 = [\bW^{d'T}\ \bw^{0T} ]^T$\\
        $\cdot$ Employ Algorithm \ref{alg:CS_BA} to obtain new $\bW^* \in \calM_{d'}^d$ and $\bc^{*} \in \R^{d'}$ while the first $d'-1$ rows of $\bW^*$ are kept fixed (and equal to $\bW^{d'})$.
}
\end{algorithm}

In all numerical examples presented in this paper, we perform the $\ell_1$-minimization problem by employing the Douglas-Rachford algorithm \cite{DR_algo_1, DR_algo_2}, that is a splitting technique of
finding a zero of the sum of two maximally monotone operators. For the
optimization with respect to the projection matrix subject to
orthogonality constraints, we make use of the Sequential Quadratic
Programming (SQP) algorithm (\cite{nocedal}, Ch. 18) that solves a
sequence of subproblems that optimize a quadratic model of the
objective function. SQP requires knowledge of the gradients of the
objective function which are derived in
\ref{sec:log_like_grad}. Another alternative for optimization problems
with orthogonal constraints would be to follow a Crank-Nicolson-like
update scheme \cite{tripathy}. However, our implementations
primarily focus on Algorithm \ref{alg:CS_BA_2} which involves
optimization with respect to one matrix row at a time, and we do not
pursue extensive exploration of more sophisticated optimization
algorithms at this time.

\section{Examples}
\label{sec:examples}

A set of numerical examples are presented below to demonstrate
the performance of our methodology. The algorithm is validated in
Sec. \ref{sec:ridge} on a synthetic example where the exact adaptation
and solutions are known. In Sec. \ref{sec:burgers}, the technique is
applied on a high-dimensional benchmark UQ problem: the stochastic
Burgers' equation with a 20-dimensional random forcing term. We
compare two cases where the first has a relatively benign random
forcing with decaying amplitudes, and the other has non-decaying
amplitudes to further challenge our algorithm.
We end this section with a
realistic engineering application involving LES of turbulent reactive
flows in a scramjet engine combustor (Sec. \ref{sec:scram}), where
engine performance QoIs are functions of 11 uncertain input
parameters, and a full-dimensional chaos expansion would be infeasible
due to the computational requirements of the simulations.

\subsection{Ridge function with known adaptation}
\label{sec:ridge}

We consider the function $u: \R^d \to \R$ that is given by
\begin{equation}
u(\bxi) = \sum_{i=1}^d \xi_i + 0.25\left(\sum_{i=1}^d \xi_i \right)^2 + 0.025 \left(\sum_{i=1}^d\xi_i\right)^3
\end{equation}
which is  a PC expansion due to its polynomial form, and the
coefficients can easily be identified. Since $\sum_{i=1}^d\xi_i$ is a zero-mean Gaussian with variance equal to $d$, the above expression can be rewritten as a function of the transformed standard Gaussian variable \begin{equation}
\label{eq:eta1}
\eta_1 = d^{-1/2} \sum_{i=1}^d\xi_i,
\end{equation}
resulting in
\begin{eqnarray}
\begin{array}{rcl} u(\eta_1) & = & d^{1/2} \eta_1 + 0.25 d \eta_1^2 + 0.025 d^{3/2} \eta_1^3 \\ & = & c_0 + c_1 \psi_1(\eta_1) + c_2 \psi_2(\eta_1) + c_3 \psi_3(\eta_1),
\end{array}
\end{eqnarray}
where
\begin{eqnarray}
\bc = \left(\begin{array}{c}c_0 \\ c_1 \\ c_2 \\ c_3 \end{array}\right) = \left(\begin{array}{c} 0.25 \\  d^{1/2} + 0.075 d^{3/2} \\ \frac{0.25d}{\sqrt{2}}\\ \frac{0.025 d^{3/2}}{\sqrt{3!}} \end{array} \right).
\end{eqnarray}

\begin{figure}[h]
\centering
\includegraphics[width = 0.49\textwidth]{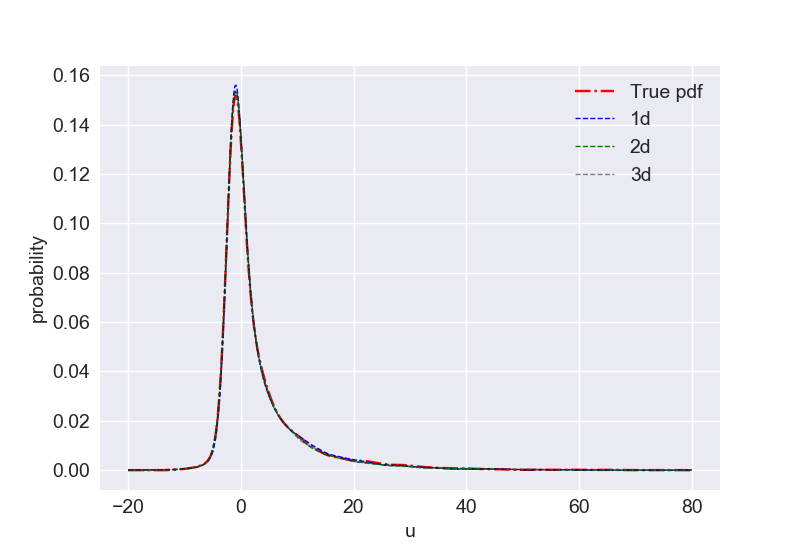}
\includegraphics[width = 0.49\textwidth]{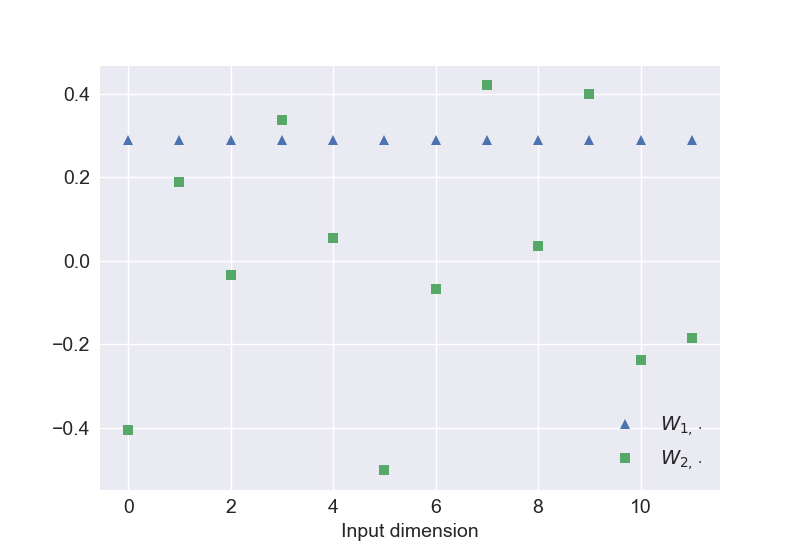}
\caption{Left: Plots of the density functions of the true QoI along with the 1-, 2- and 3-dimensional adapted chaos approximations. Right: Values of the first two rows ($\bW_{1,\cdot}$ and $\bW_{2,\cdot}$) of the projection matrix, estimated using Alg. \ref{alg:CS_BA_2}.\label{fig:pdf_u_plot}
}
\end{figure}

\begin{figure}[h]
\centering
\includegraphics[width = 0.49\textwidth]{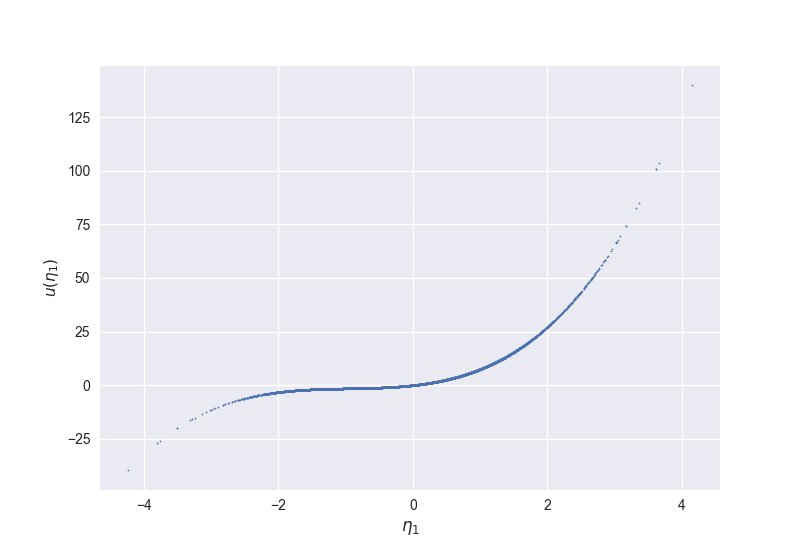}
\includegraphics[width = 0.49\textwidth]{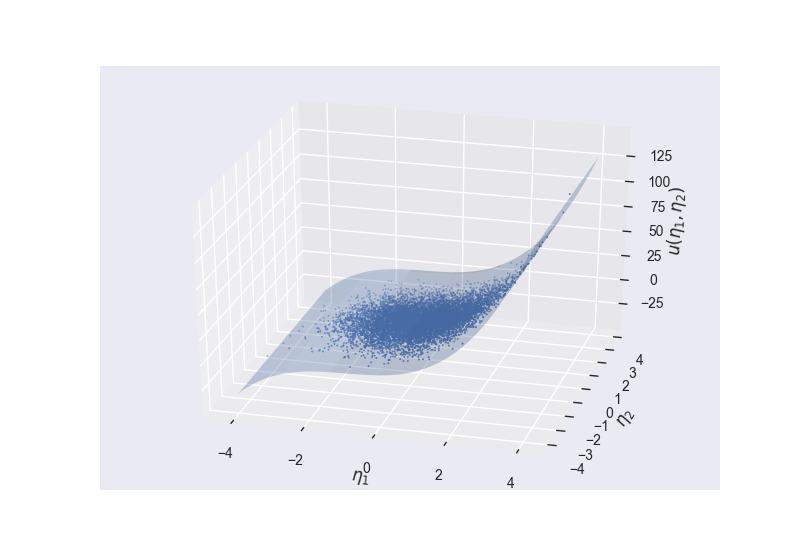}
\caption{Plot of the QoI as a function of its 1d and 2d inputs (left and right respectively). \label{fig:toy_u_plot}}
\end{figure}

\begin{figure}[h]
\centering
\includegraphics[width = 0.49\textwidth]{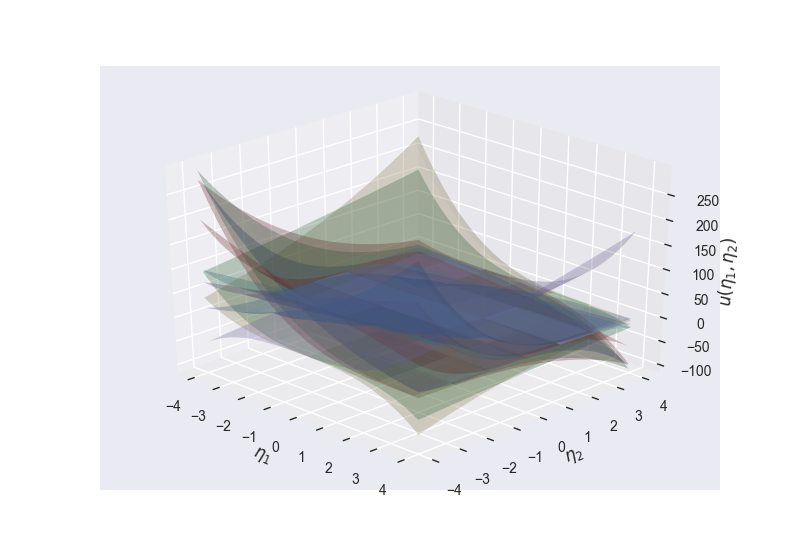}
\includegraphics[width = 0.49\textwidth]{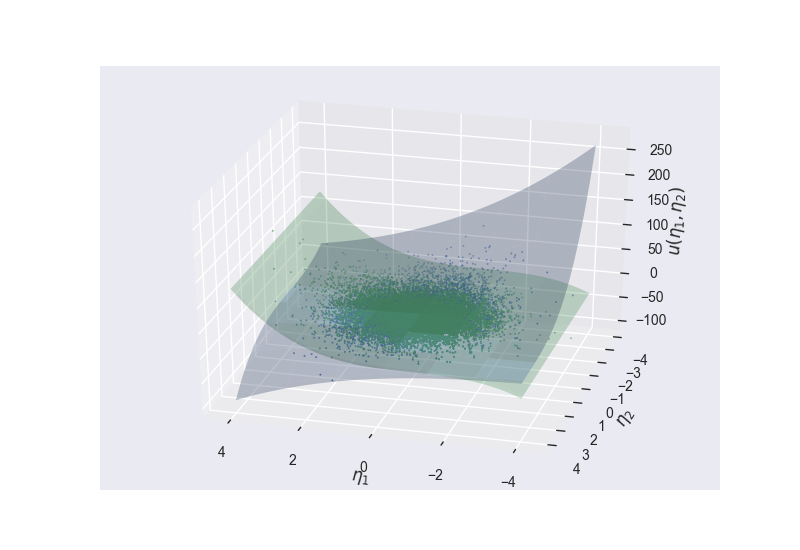}
\caption{Left: Bivariate PC expansions obtained from $10$ runs of Algorithm \ref{alg:CS_BA} with random initial point. Right: Comparison of a single run with the output of Algorithm \ref{alg:CS_BA_2}.\label{fig:toy_alg1}}
\end{figure}

We set $d = 12$, therefore $|\calJ_3^{12}| = 455$ and we construct
synthetic data that consists of $N = 180$ Monte Carlo samples ($N /
|\calJ_3^{12}| \approx 0.4$). We execute Algorithm \ref{alg:CS_BA_2}
and obtain the solutions for 1d, 2d, and 3d
expansions. Fig. \ref{fig:pdf_u_plot} shows all three density
functions (left) and isometry values for the 1d and 2d cases (right). It is clear
that the densities coincide since $u(\bxi)$ can be written as a
univariate function. The first row of the isometry is indeed as in
(\ref{eq:eta1}) while the values of the second row are in fact insignificant since the series coefficients
that correspond to $\eta_2$ (and cross terms) are zero.                 Fig. \ref{fig:toy_u_plot} shows the plot of $u$ as
a function of $\eta_1$ (left) and as a bivariate function of $(\eta_1,
\eta_2)$ (right). Since the coefficients corresponding to $\eta_2$ are
zero, the function exhibits no variation along
$\eta_2$. Fig. \ref{fig:toy_alg1} shows the bivariate (2d) expansions
obtained after performing 10 independent runs of Algorithm
\ref{alg:CS_BA} and a comparison of one run from each
algorithm. Interestingly we observe that at each run, Algorithm
\ref{alg:CS_BA} converges to an arbitrary rotation and the
corresponding coefficients result in an expansion $u(\eta_1',
\eta_2')$ that itself is a rotation of $u(\eta_1, \eta_2)$ obtained by
Algorithm \ref{alg:CS_BA_2}. That is due to the fact that the
observations incorporated in the $\ell_2$ error term are each time
mapped to different rotated inputs. Thus both algorithms capture the
same PC expansion but Alg. \ref{alg:CS_BA} fails to detect a dominant
direction.

\subsection{Stochastic Burgers' equation}
\label{sec:burgers}

Let us consider the following initial boundary value problem (IBVP):
\begin{eqnarray}
\label{eq:burgers}
\left\{
\begin{array}{l}
\frac{\partial v}{\partial t} + v \frac{\partial v}{\partial x} = \nu \frac{\partial^2 v}{\partial x^2} + \sigma \sum_{l=1}^M\xi_l \phi_l(x,t), \ \ x \in [0, 2\pi], \ \ t\in[0, 1]\\
\\
v(x, 0) = 1 + \sin(2x)\\
\\
v(0, t) = v(2\pi, t) = 1 + \sin(\pi t)
\end{array}\right.
\end{eqnarray}
where $\xi_l$, $l = 1,\dots, M$ are i.i.d. standard normal random variables. For the random forcing term we consider two cases: (i) $\phi_l(x,t) = \cos(2l x)\cos(2l\pi t) / \sqrt{l}$ where the strength of the random Gaussians decays as a function of $l$ and (ii) $\psi_l(x,t) = \cos(2l x)\cos(2l\pi t) / M$ where all terms contribute equally although their coefficients maintain varying (increasing) frequencies. For our numerical implementations below we discretize $[0,2\pi]\times [0,1]$ into a rectangular $500\times 500$ grid and solve the IBVP using an implicit Newton's method. The scalar QoI which we seek to expand in a PC series with respect to $\bxi = (\xi_1,\dots, \xi_M)$ is the spatial average of the solution to the IBVP at $t = 1$,\begin{equation}
u(\bxi) := \frac{1}{2\pi}\int_0^{2\pi} v(x,1;\bxi) dx.
\end{equation}
For both cases we take $M=20$ with $\nu = 1/2$ and $\sigma = 2$. For case (i) we set the order of the approximating expansion expansion to be $Q = 6$ and we generate $1000$ Monte Carlo samples as our synthetic data while for (ii) we reduce the order to $Q = 3$ and the number of samples is set to $700$.

\begin{figure}
\centering
\includegraphics[width = 0.49\textwidth]{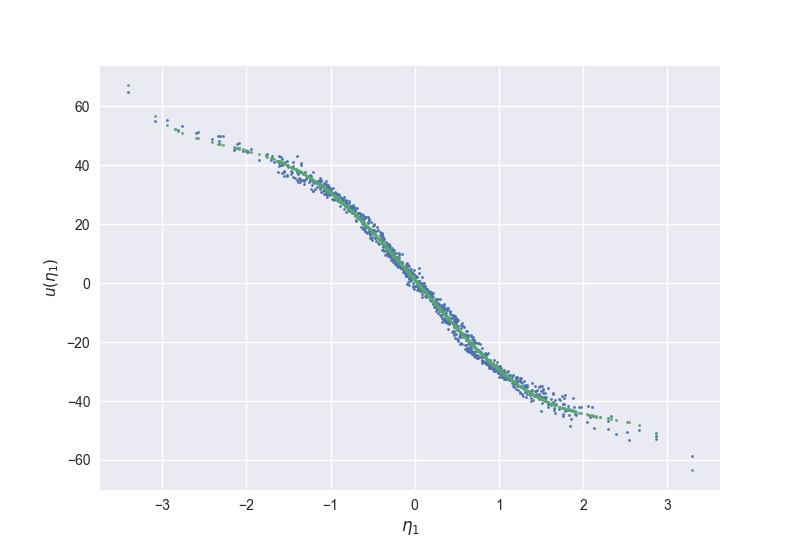}
\includegraphics[width = 0.49\textwidth]{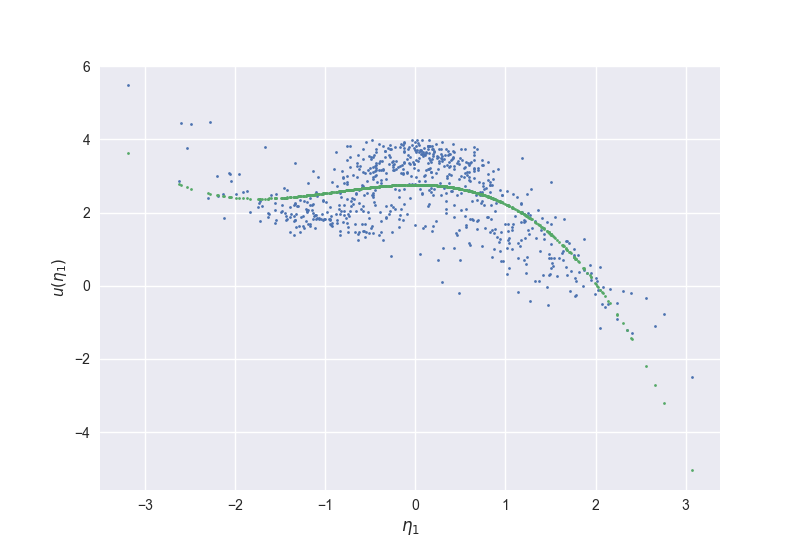}\\
\includegraphics[width = 0.49\textwidth]{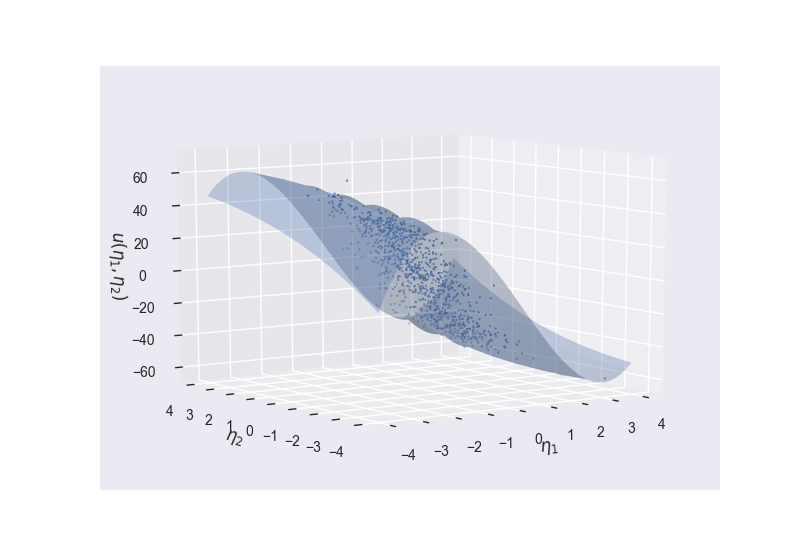}
\includegraphics[width = 0.49\textwidth]{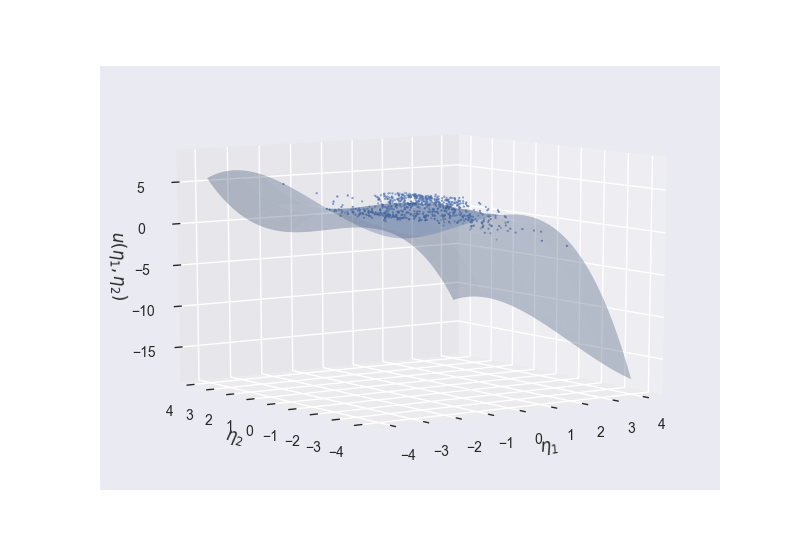}\\
\includegraphics[width = 0.49\textwidth]{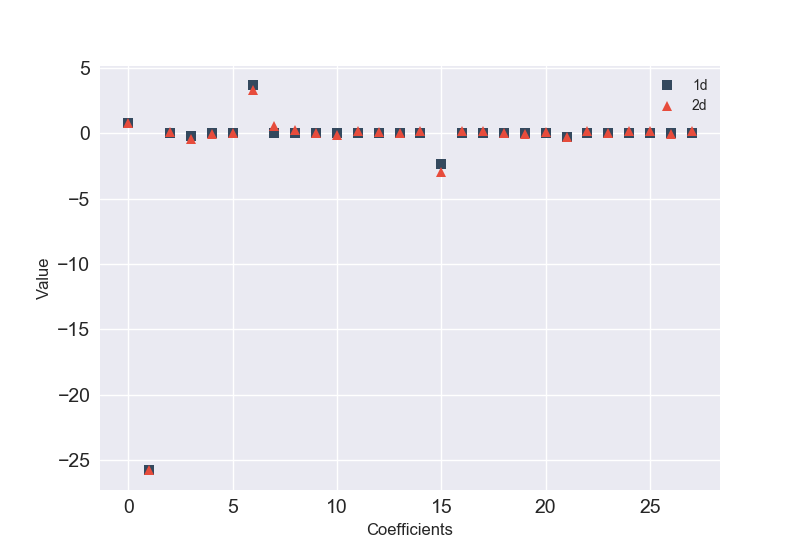}
\includegraphics[width = 0.49\textwidth]{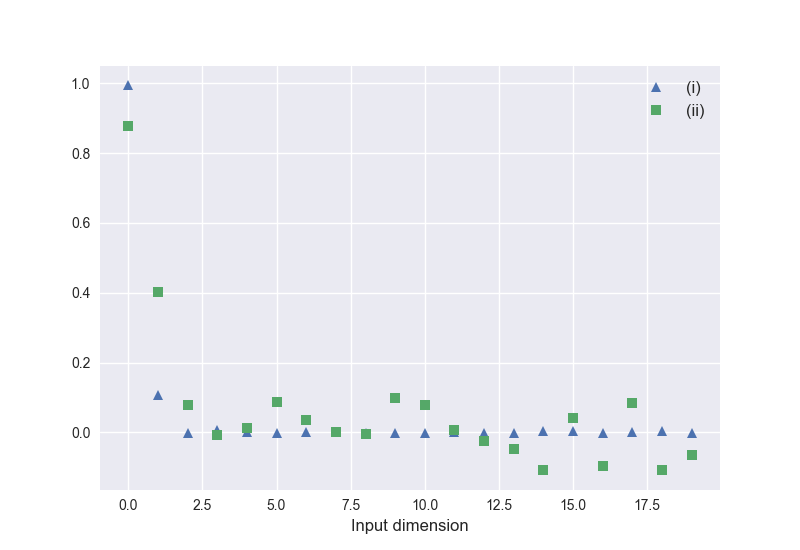}
\caption{Top: Plot of the QoI as a function of its 1d input for case (i) (left) and case (ii) (right). Middle: Plot of the QoI as a function of its 2d input for case (i) (left) and case (ii) (right). Bottom: Coefficients of 1d and 2d expansions for case (i) (left) and comparison of the first row of $\bW$ for the two cases (right). \label{fig:burgers_u_plot}}
\end{figure}

\begin{figure}
\centering
\includegraphics[width = 0.49\textwidth]{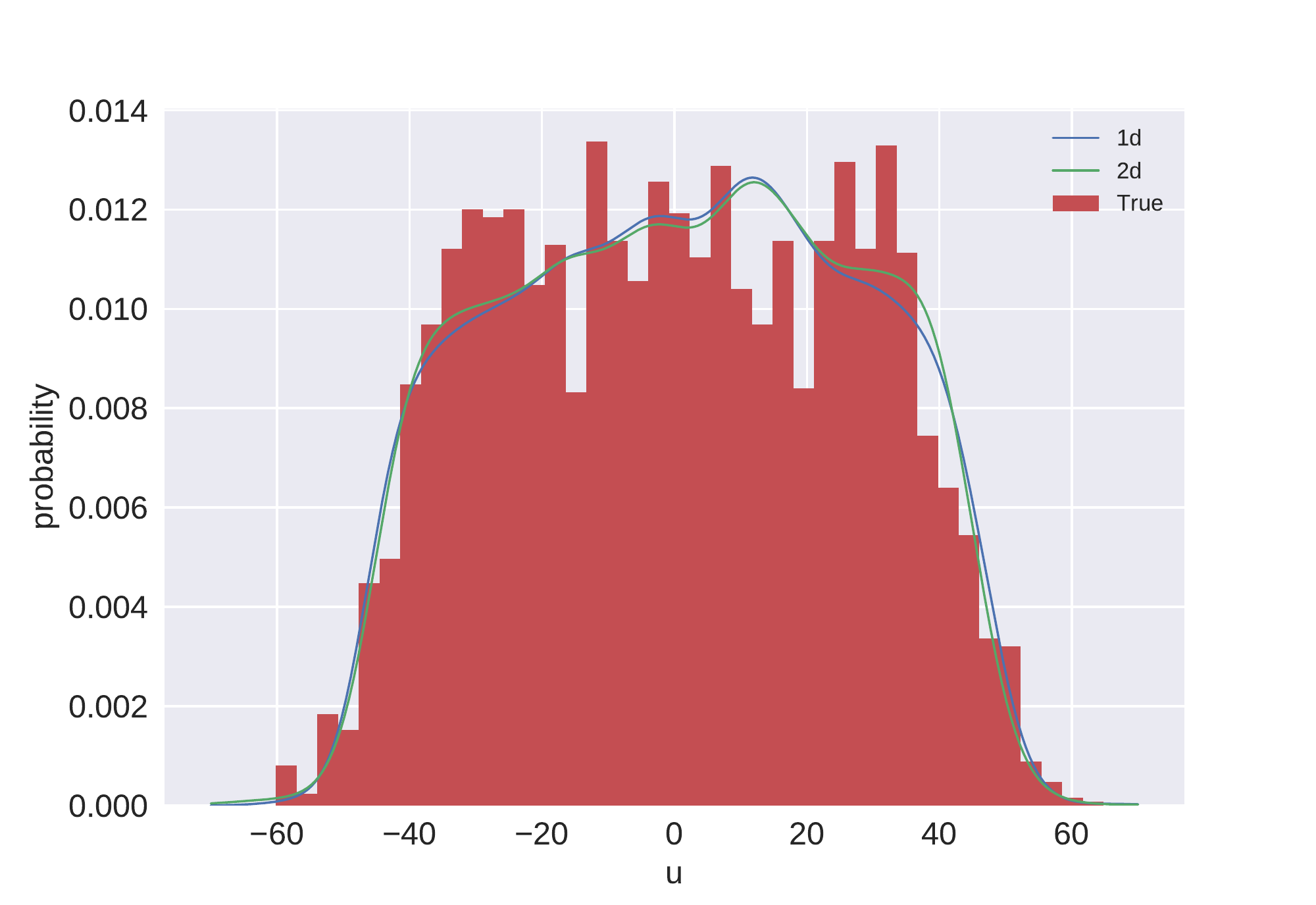}
\includegraphics[width = 0.49\textwidth]{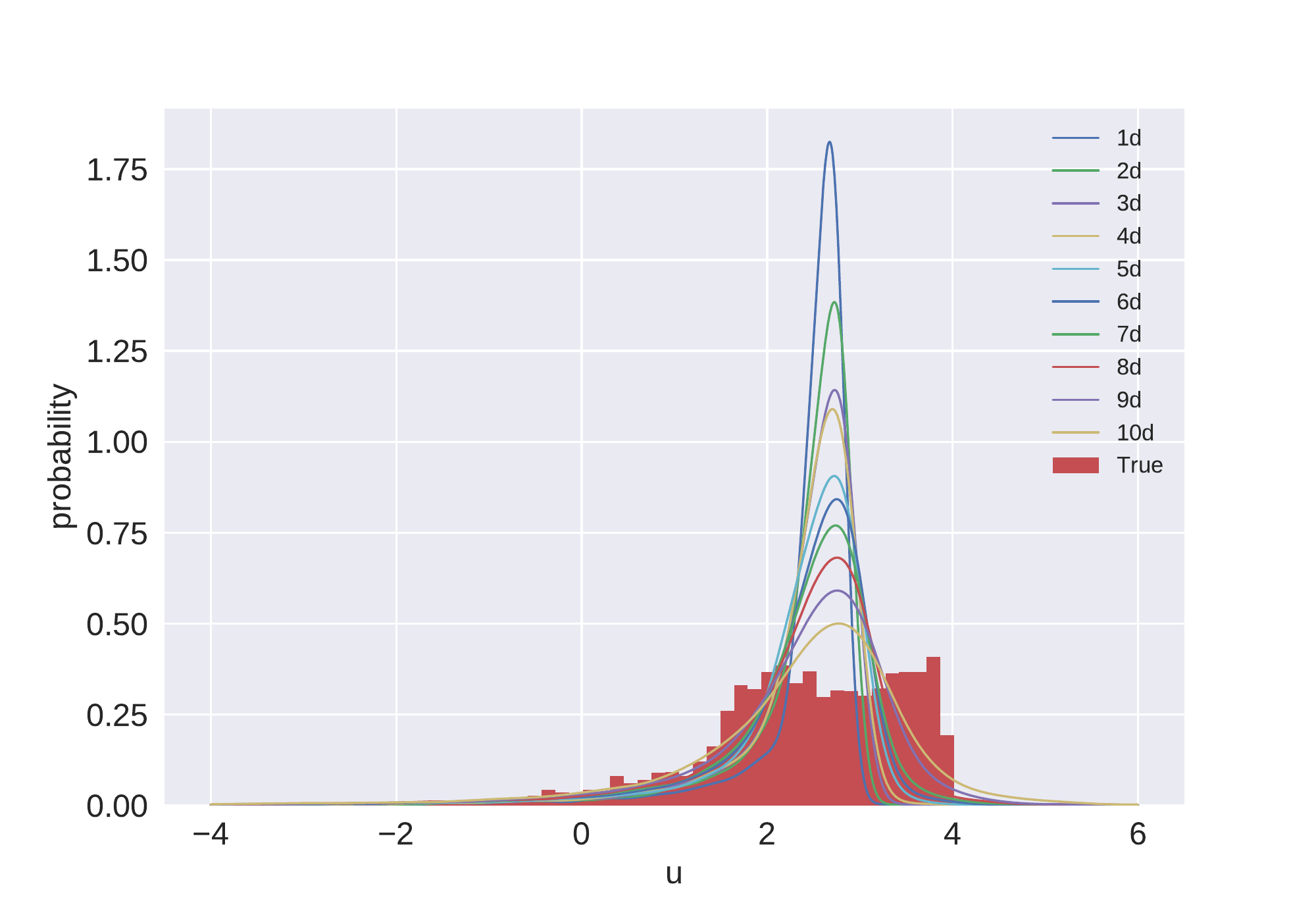}
\caption{Density functions of the PC expansions for $d_0 = 1, 2$ in case (i) (left) and for $d_0 = 1,\dots, 10$ in case (ii) (right). To allow comparison with the true pdf of each QoI, we also display the histograms that correspond to an empirical distribution and are based on $5000$ MC samples drawn by solving eq. (\ref{eq:burgers}) directly. \label{fig:burgers_pdfs}}
\end{figure}

\begin{figure}
\centering
\includegraphics[width = 0.32\textwidth]{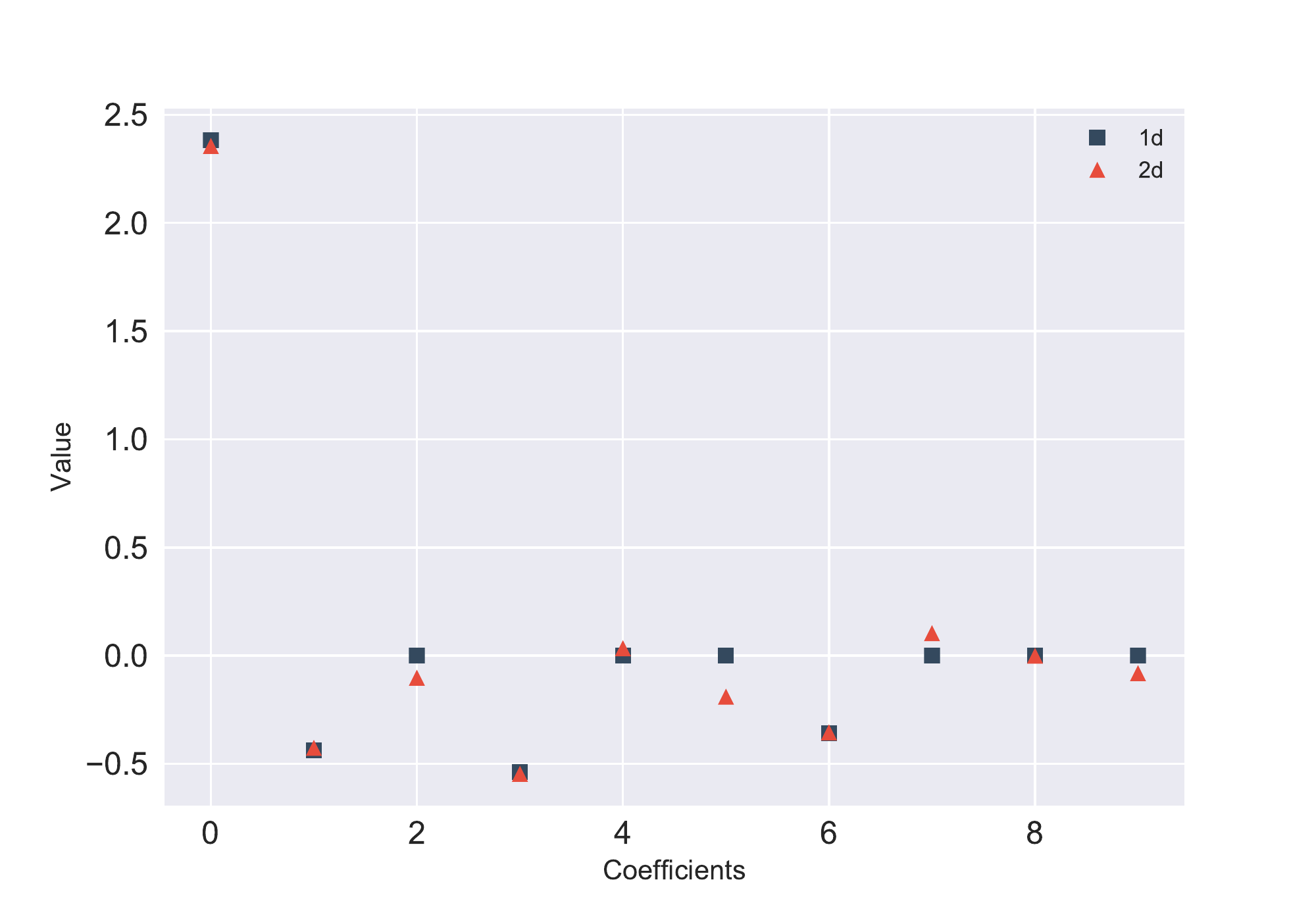}
\includegraphics[width = 0.32\textwidth]{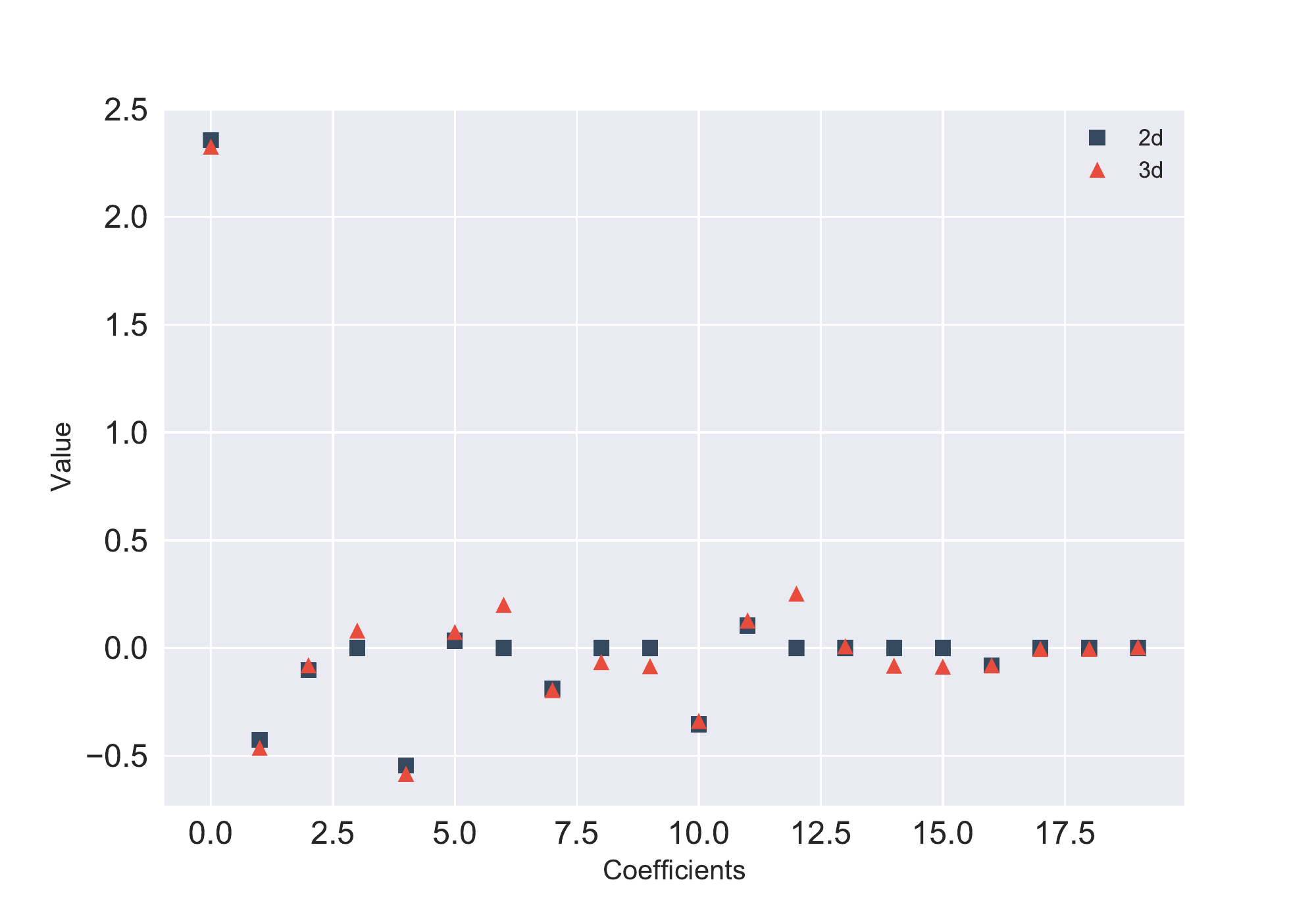}
\includegraphics[width = 0.32\textwidth]{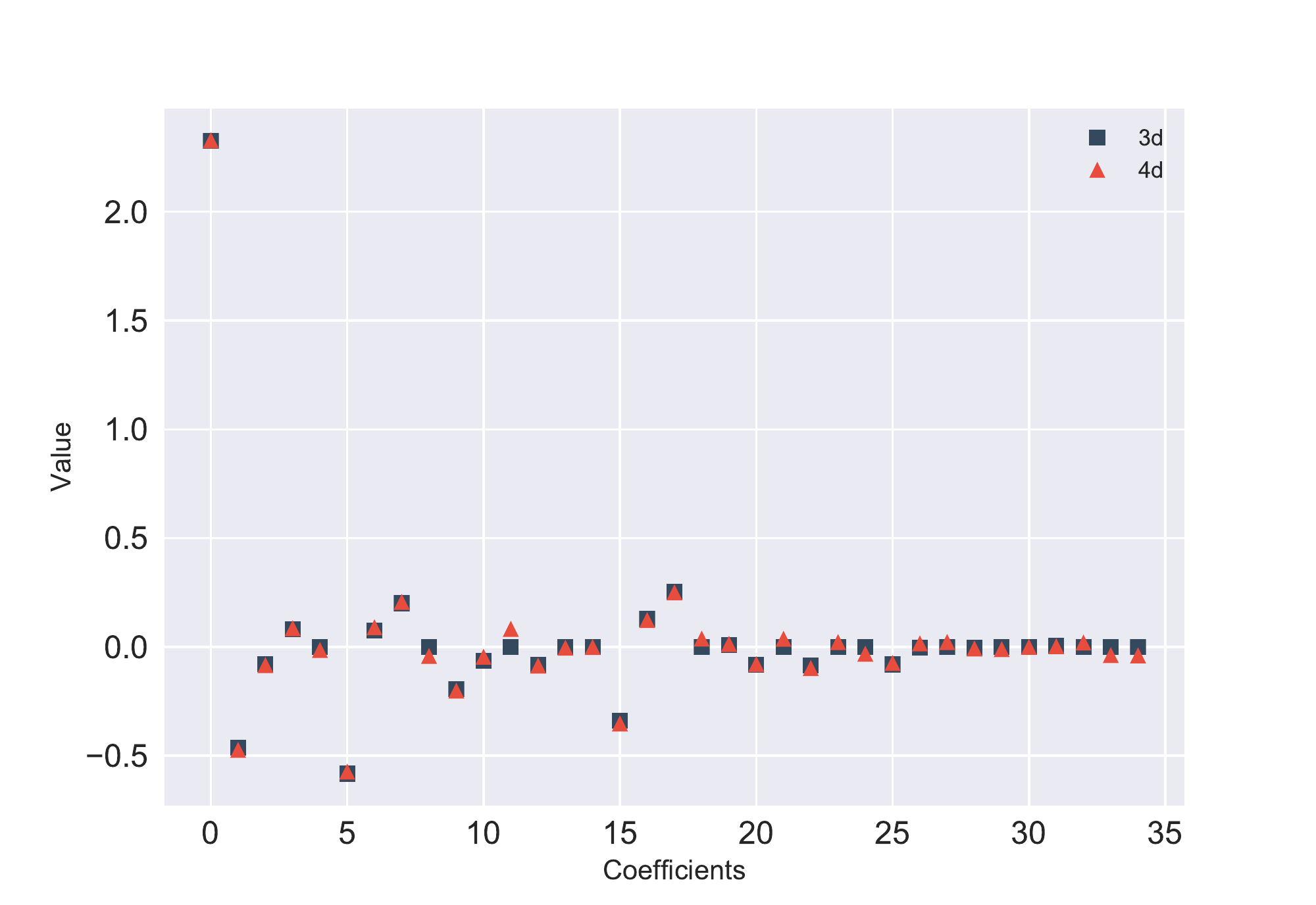}
\includegraphics[width = 0.32\textwidth]{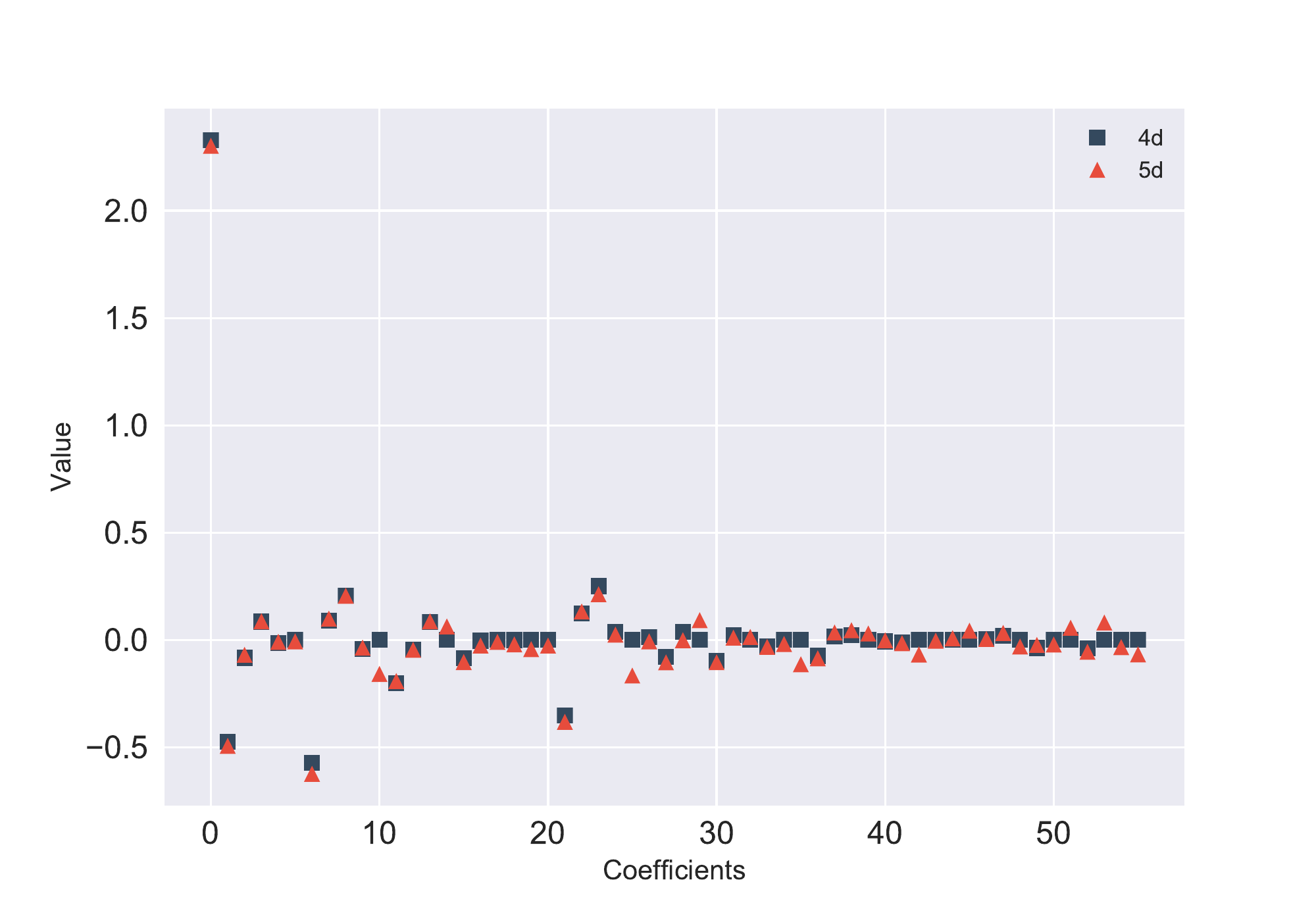}
\includegraphics[width = 0.32\textwidth]{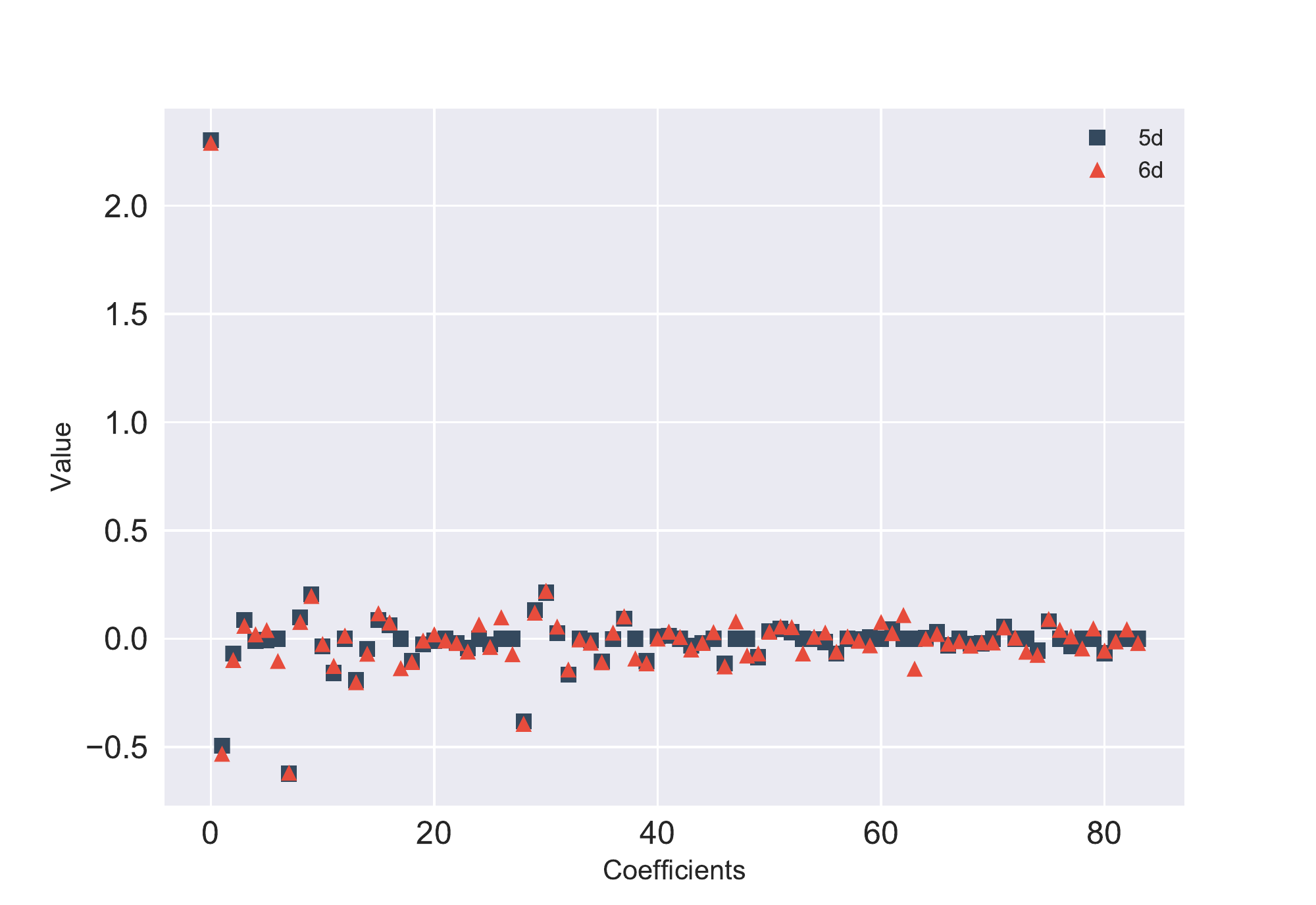}
\includegraphics[width = 0.32\textwidth]{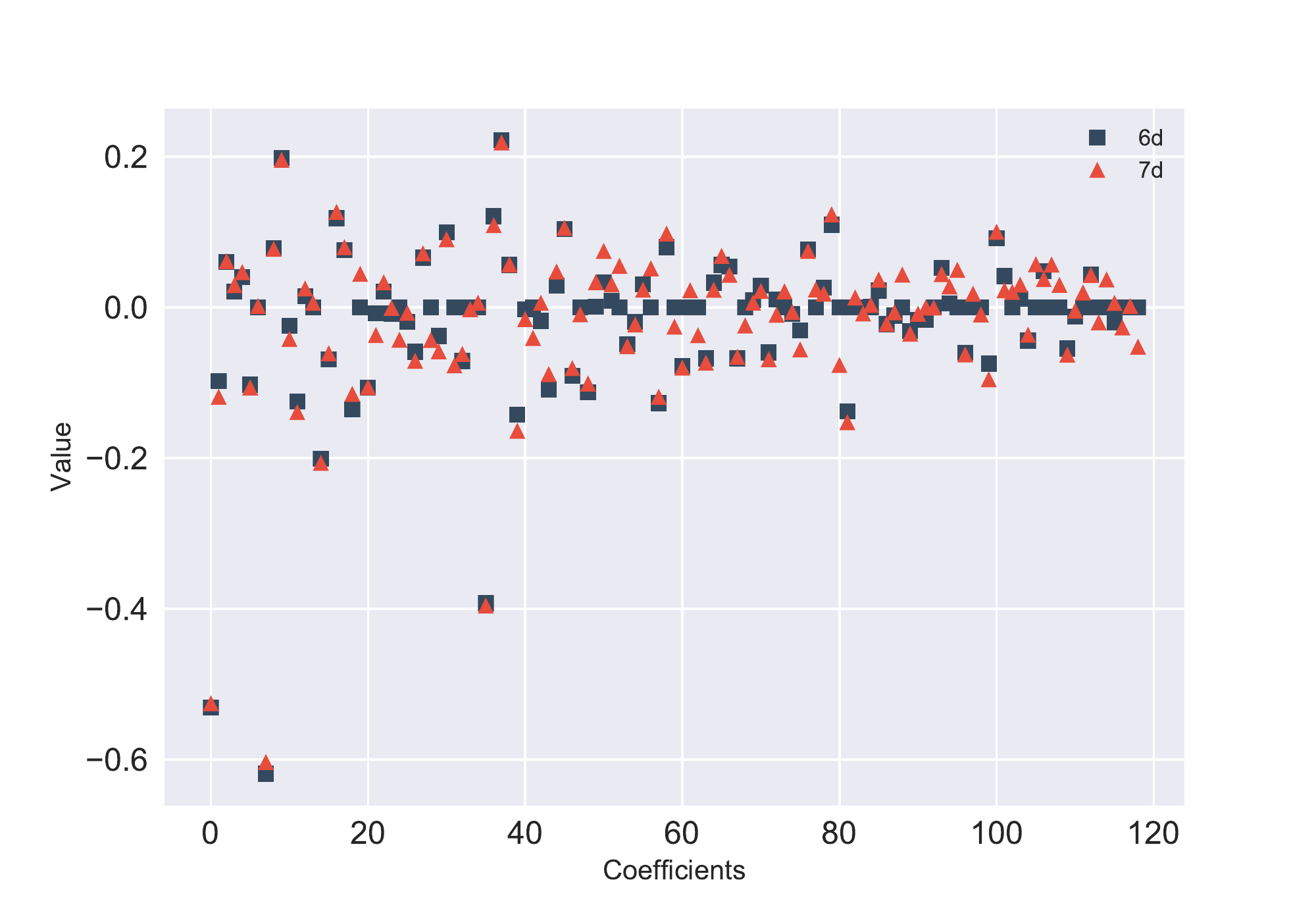}
\includegraphics[width = 0.32\textwidth]{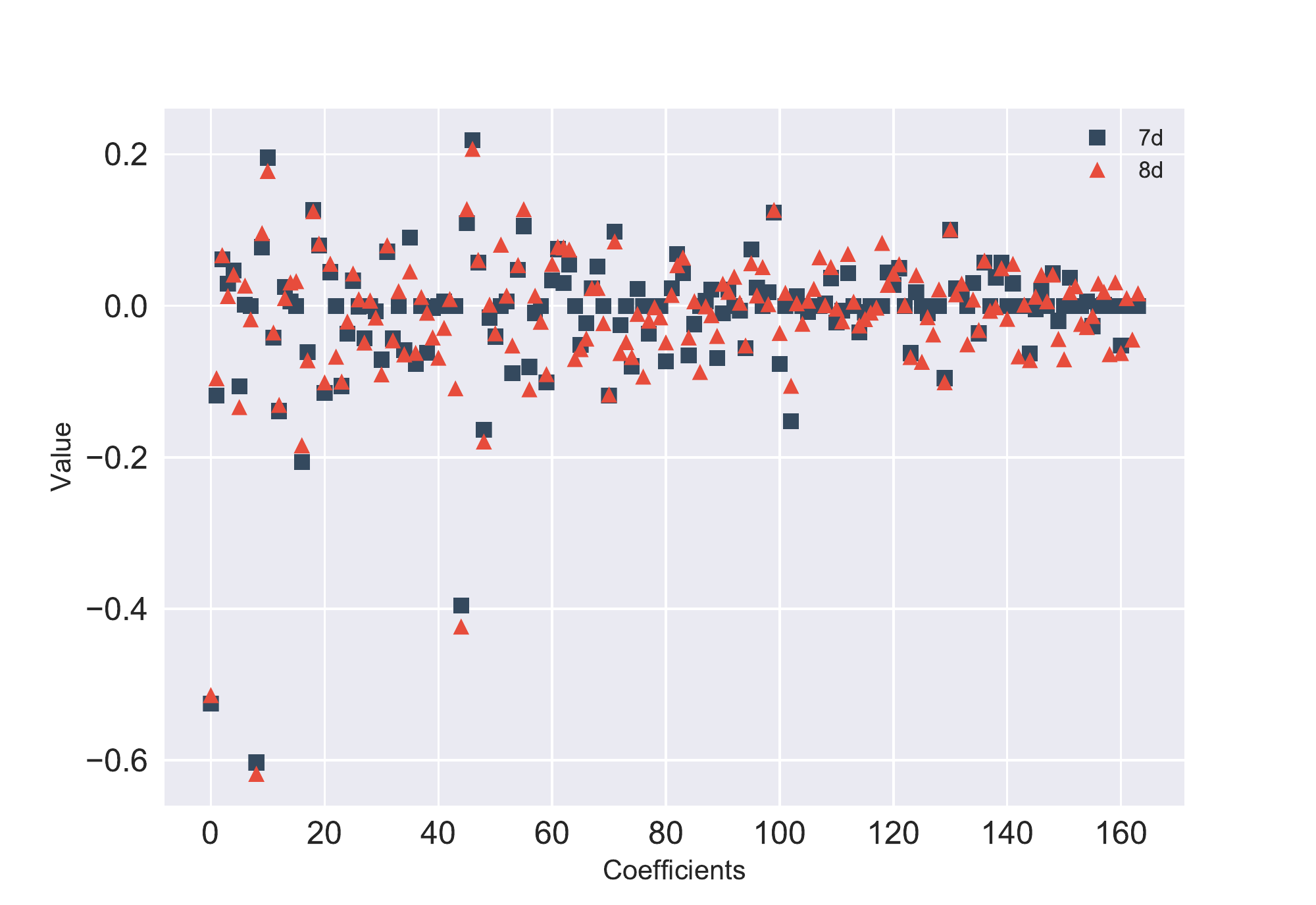}
\includegraphics[width = 0.32\textwidth]{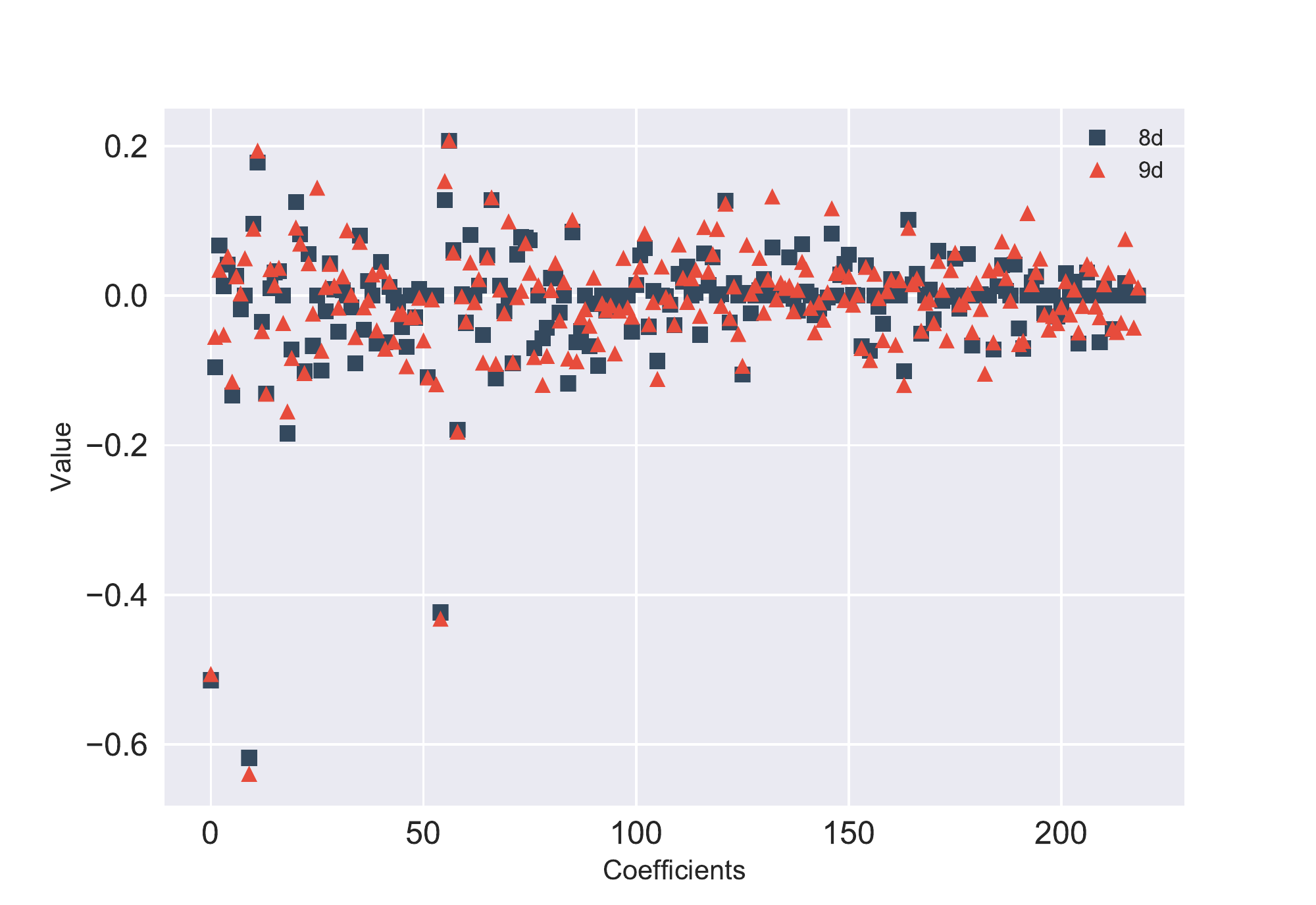}
\includegraphics[width = 0.32\textwidth]{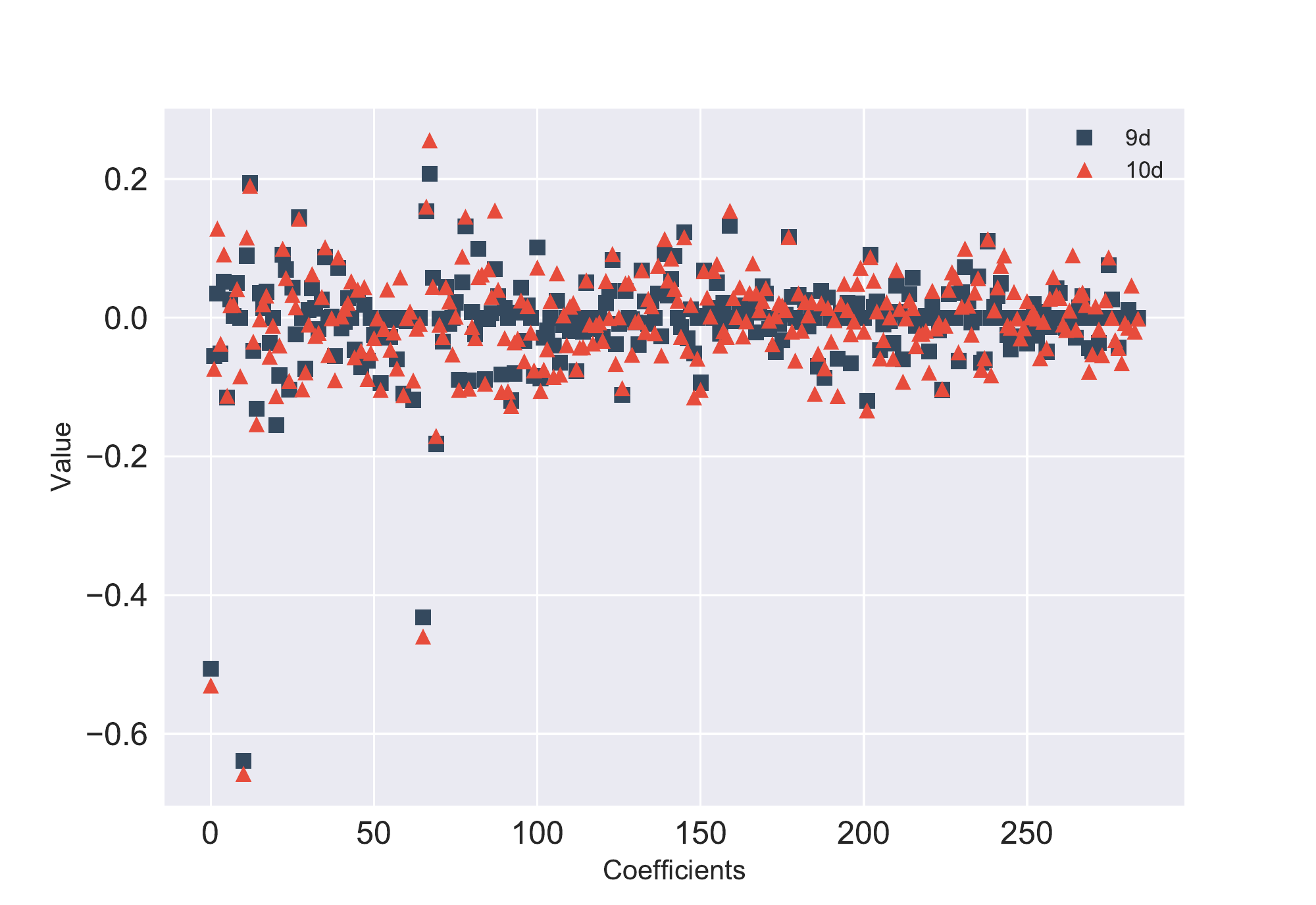}
\caption{Case (ii): Plot of chaos coefficients for expansions with consecutive increase in dimensionality, from 1d-2d (top left) to 9d-10d (bottom right). To improve visibility along the $y$-axis, the zeroth order coefficient is ignored in the comparison of 6d-7d and higher orders.\label{fig:burgers_coeffs}}
\end{figure}

Fig. \ref{fig:burgers_u_plot} shows the plots of the estimated 1d and 2d expansions obtained from Algorithm \ref{alg:CS_BA_2} for the two cases. For case (i), the expansions provide a good fit the data. This can be partially explained by the fact that $Q$ is higher but most importantly because the decay in the random forcing proportional to $1/\sqrt{l}$ quickly makes $\xi_i$'s insignificant, and the QoI depends on only a few inputs, thus making it easy to identify a rotation in a low dimensional space. On the contrary, in case (ii) clearly both expansions provide a quite poor fit on the data and the need to increase $d_0$ is apparent. A comparison of the coefficients of the 1d and 2d expansions for (i) indicates that the polynomial terms that depend on $\eta_1$ are dominant compared to those that depend on $\eta_2$. In addition, a look at the entries of the first row of the projection matrix for the two cases confirms our assumption above regarding the increasing significance of the $\xi_i$'s as we move from case (i) to case (ii). In the first, only two entries have significant amplitudes, while in the second, the values exhibit fluctuations that result in $\eta_1$ being strongly dependent on all $\xi_i$'s.
 Fig. \ref{fig:burgers_pdfs} shows the density functions of the PC expansions
 obtained for the two cases. For (i) we compare the PC expansions of dimension up
 to $2$ as we find no reason to pursue estimation of higher dimensional
 expansions. For (ii) we display the densities for expansions of dimensionality
 up to $10$ where it is observed that no convergence is yet achieved. For comparison, we also display the empirical densities (histograms) of the QoI based on $5000$ samples that are drawn by directly solving eq. (\ref{eq:burgers}). The
 coefficients of all successive PC expansions are pairwise compared in Fig.
 \ref{fig:burgers_coeffs}. We observe that by using the computed rotation found
 for $d'-1$, each successive run of the algorithm for $d'$ seems to recover the
 coefficients of the $d'-1$ augmented with the additional nonzero coefficients
 corresponding to $\eta_{d'}$ plus cross terms. Overall, the method performs effectively in both cases. However, only in the first case we manage to obtain a reduced PC expansion that can be used as an approximation of our QoI. This is due to the different effect on the random forcing on the QoI and independent of our algorithm. Nevertheless, for the second and more challenging case we still manage to draw our conclusions at a fixed computation cost, that of performing $700$ runs of the PDE solver, in constrast to using quadrature methods which would require far more model evaluations. For instance, using a level 1 quadrature rule to compute first order coefficients as in \cite{tipireddy}, and then using a level 3 (to account for Q = 3 in this case) quadrature rule on the reduced basis from $d_0 = 1$ up to $10$, would require a total of $8501$ evaluations!   

\subsection{Turbulent reactive flows in a scramjet engine combustor}
\label{sec:scram}

UQ for supersonic reactive flows using large eddy simulations (LES)
has only recently become feasible owing to both algorithmic advances
and increasing computational power and resources. This development has
allowed researchers to explore beyond the commonly used
Reynolds-averaged Navier-Stokes (RANS) model \cite{yentsch}.  Even
with the use of RANS, hybrid RANS/LES, or Detached Eddy simulations
(DES) \cite{spalart}, construction of accurate response surfaces for
QoIs faces insurmountable challenges due to the large number of
simulations required to explore the often high-dimensional space of
uncertain model parameters.  Indeed, systematic UQ studies for
supersonic combusting ramjet (scramjet) engines is currently rare,
with a few exceptions \cite{Witteveen2011, constantine_hyshot}.
Only very recently, CS methods were used for constructing PC
surrogates for scramjet computations \cite{huan_CS} and global
sensitivity analysis studies were presented \cite{huan_sobol}.

\subsubsection{The model}

\begin{figure}[hbt]
  \centering
  \includegraphics[width=0.6\textwidth]{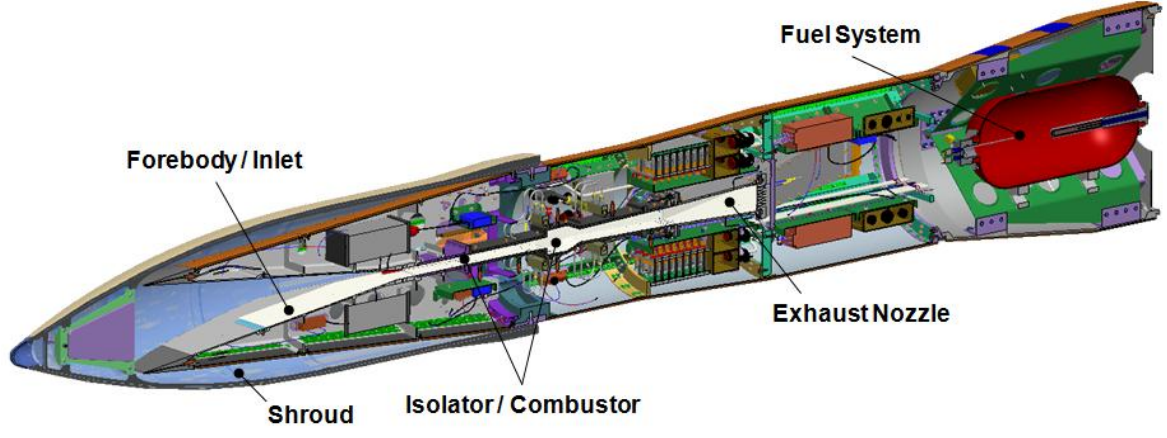}
  \includegraphics[width=0.35\textwidth]{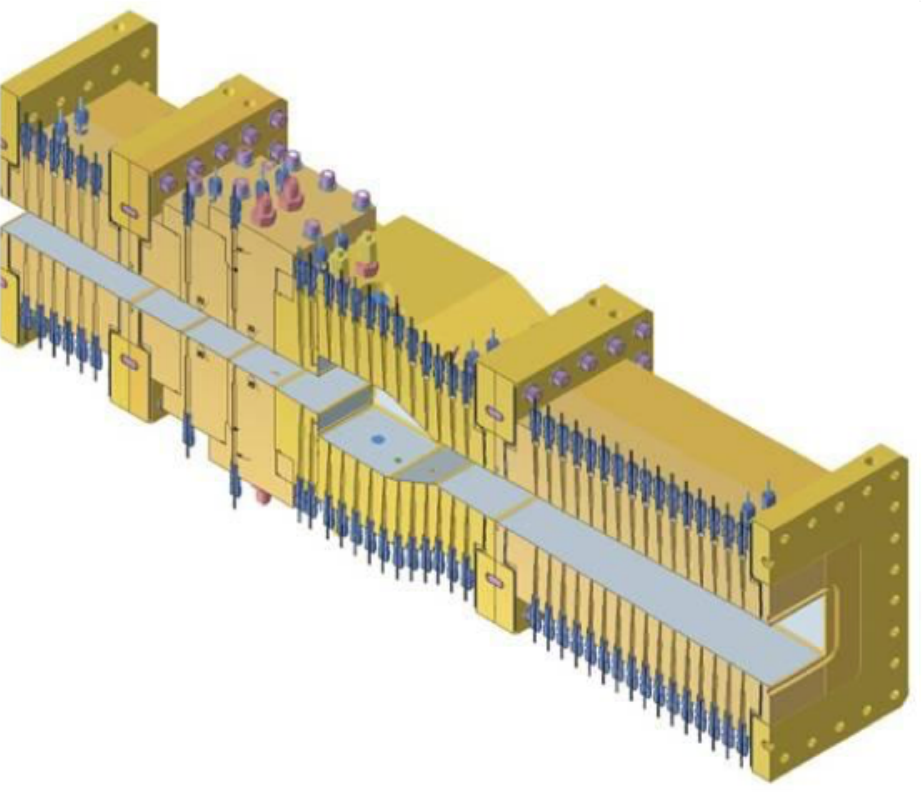}
  \caption{HIFiRE Flight 2 payload (left) \cite{jackson} and HDCR cut view (right) \cite{hass}.                 \label{f:HIFiRE}}
\end{figure}

We concentrate on a scramjet configuration studied under the HIFiRE
(Hypersonic International Flight Research and Experimentation)
program~\cite{Dolvin2008, Dolvin2009}, where the flight test payload
(Figure \ref{f:HIFiRE}) involves a cavity-based hydrocarbon-fueled dual-mode
scramjet.
A ground test rig, designated the HIFiRE Direct Connect Rig (HDCR)
(Figure \ref{f:HIFiRE}), 
was developed to duplicate the isolator/combustor
layout of the flight test hardware~\cite{hass, Storch2011}. Mirroring
the HDCR setup, we aim to simulate and assess flow characteristics
inside the isolator/combustor portion of the scramjet.

\begin{figure}
  \centering
  \includegraphics[width=0.7\textwidth]{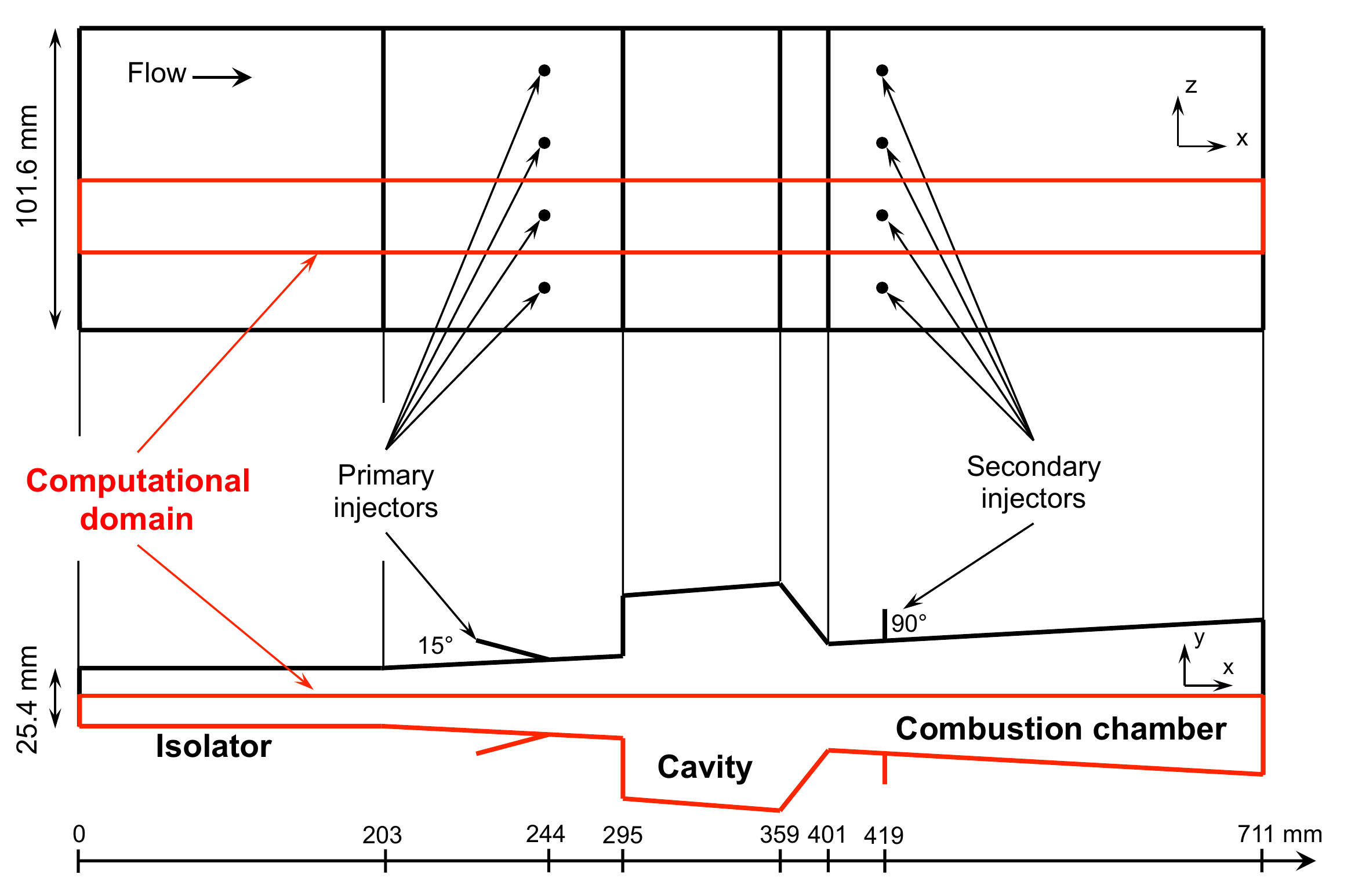}
  \caption{The HDCR experimental setup and schematic of the full
  computational domain \cite{Lacaze2017}.     \label{f:full_schematic}}
\end{figure}

We simulate reactive flow through the HDCR. The rig consists of a
constant-area isolator (planar duct) attached to a combustion chamber.
It includes four primary injectors that are mounted upstream of flame
stabilization cavities on both the top and bottom walls. Four
secondary injectors along both walls are positioned downstream of the
cavities.  The primary fuel injectors are located at $x = 244$ mm from
the inlet and aligned at $15^\circ$ from the wall, while the secondary
injectors are at $x = 419$ mm and aligned at $90^\circ$ from the
wall. All injectors have a diameter of $d = 3.175$ mm.  Flow travels
from left to right in the $x$-direction (streamwise), and the geometry
is symmetric about the centerline in the $y$-direction.  Numerical
simulations take advantage of this symmetry by considering a domain
that covers only the bottom half of this configuration. To further
reduce the computational cost, we consider one set of
primary/secondary injectors and impose periodic conditions in the
$z$-direction (spanwise).  The overall computational domain is
highlighted by the red lines in Figure \ref{f:full_schematic}.
JP-7 surrogate fuel~\cite{Pellett2007a} is inserted through these
injectors, containing 36\% methane and 64\% ethylene by volume. A reduced, three-step mechanism is employed to characterize
the combustion process, and its kinetic parameters are tuned for the
current simulations~\cite{Lacaze2017}.

LES calculations are then performed using the RAPTOR code framework
developed by Oefelein~\cite{oefelein, oefelein_phd}. The solver has
been optimized to meet the strict algorithmic requirements imposed by
the LES formalism. The theoretical framework solves the fully coupled
conservation equations of mass, momentum, total-energy, and species
for a chemically reacting flow. It is designed to handle high Reynolds
number, high-pressure, real-gas and/or liquid conditions over a wide
Mach operating range. It also accounts for detailed thermodynamics and
transport processes at the molecular level. RAPTOR employs
non-dissipative, discretely conservative, staggered, finite-volume
differencing, which eliminates numerical contamination due to
artificial dissipation and produces high quality LES results.

\subsubsection{Input parameters and quantities of interest}

In our example, we allow a total of 11 input parameters to be variable
and uncertain, shown in Table \ref{tab:params} along with the range of
admissible values. Their distributions are assumed uniform across the
ranges indicated and, for the purpose of constructing a Hermite Chaos
expansion, are further mapped to Gaussian variables as explained in
the next section.  We focus on two QoIs: (1)
burned equivalence ratio ($\phi_{B}$) and (2) stagnation pressure loss
ratio ($R_{\bar{P}}$).
These QoIs reflect the overall scramjet performance, and are based on
time-averaged variables. The data utilized in the current analysis are
from 2D simulations on the $(x,y)$ plane of the scramjet computational domain (bottom of Fig. \ref{f:full_schematic}), using grid
resolution where cell size is $1/16$ of the injector
diameter $d=3.175$ mm.
\begin{itemize}
\item \textbf{Burned equivalence ratio ($\phi_{B}$)} is defined to be
  equal to $\phi_{B} \equiv \phi_{T} \eta_{c}$, where $\phi_{T}$ is
  the total equivalence ratio imposed on the system, and $\eta_{c}$ is
  the combustion efficiency based on static enthalpy
  quantities~\cite{Storch2011, Gruber2008}:
  \begin{align}
    \eta_{c} =
    \frac{H(\Tref,\Yexit)-H(\Tref,\Yin)}{H(\Tref,\Yideal)-H(\Tref,\Yin)}.\label{e:eta_c}
  \end{align}
  Here $H$ is the total static enthalpy, the ``ref'' subscript
  indicates a reference condition derived from the inputs, the ``e''
  subscript is for the exit, and the ``ideal'' subscript is for the
  ideal condition where all fuel is burnt to completion.  The
  reference condition corresponds to that of a hypothetical
  non-reacting mixture of all inlet air and fuel at thermal
  equilibrium. The numerator, $H(\Tref,\Yexit)-H(\Tref,\Yin)$, thus
  reflects the global heat released during the combustion, while the
  denominator represents the total heat release available in the
  fuel-air mixture.

\item \textbf{Stagnation pressure loss ratio ($R_{\bar{P}}$)} is defined as
  \begin{align}
    R_{\bar{P}} = 1 - \frac{P_{s, e}}{P_{s, i}},
  \end{align}
  where $P_{s, e}$ and $P_{s, i}$ are the wall-normal-averaged
  stagnation pressure quantities at the exit and inlet planes,
  respectively.
\end{itemize}

\begin{table}[h]
\caption{Uncertain model parameters. The uncertain distributions are
  assumed uniform across the ranges shown.}
\label{tab:params}
\centering
\begin{tabular}{l | c | l}
 & Notation & Range \\
\hline
\hline
\emph{Inlet boundary conditions} & & \\
\hline
Stagnation pressure & $p_0$ & $[1.406, 1.554] \times 10^6$ Pa \\
Stagnation temperature & $T_0$ & $[1472.5, 1627.5]$ K \\
Mach number & $M_0$ & $[2.259, 2.761]$ \\
Turbulence intensity horizontal component & $I_i$ & $[0, 0.05]$ \\
Turbulence length scale & $L_i$ & $[0, 8] \times 10^{-3}$ m \\
Ratio of turbulence intensity vertical to horizontal components & $R_i$ & $[0.8, 1.2]$ \\
\hline
\hline
\emph{Fuel inflow boundary conditions} & & \\
\hline
Turbulence intensity magnitude & $I_f$ & $[0, 0.05]$ \\
Turbulence length scale & $L_f$ & $[0, 1]\times 10^{-3}$ m\\
\hline
\hline
\emph{Turbulence model parameters} & & \\
\hline
Modified Smagorinsky constant & $C_R$ & $[0.01, 0.06]$ \\
Turbulent Prandtl number & $Pr_t$ & $[0.5, 1.7]$ \\
Turbulent Schmidt number & $Sc_t$ & $[0.5, 1.7]$
\end{tabular}
\end{table}

\subsubsection{Results}

In order to construct Hermite Chaos expansions, we first introduce the
normalized physical parameters $\btheta = (\theta_1, \dots,
\theta_{11}) := \left( \frac{\overline{p_0}}{|p_0|},
\frac{\overline{T_0}}{|T_0|}, \frac{\overline{M_0}}{|M_0|},
\frac{\overline{C_R}}{|C_R|}, \frac{\overline{Pr_t}}{|Pr_t|},
\frac{\overline{Sc_t}}{|Sc_t|}, \frac{\overline{I_i}}{|I_i|},
\frac{\overline{L_i}}{|L_i|}, \frac{\overline{R_i}}{|R_i|},
\frac{\overline{I_f}}{|I_f|}, \frac{\overline{L_f}}{|L_f|}\right)$ where
$|\cdot|$ denotes the range of each parameter as that is shown in
Table \ref{tab:params} and the bar denotes that the parameters are
shifted towards zero (lower bound value is subtracted), hence all
parameters are normalized to $\theta_i \in [0,1]$. Next $\btheta$ is
mapped to Gaussian random germs $\bxi = (\xi_1, \dots, \xi_{11})$ via
the relation $\theta_i = \Phi(\xi_i)$, $i=1, \dots, 11$ where
$\Phi(\cdot)$ is the standard normal cumulative distribution function.

\begin{figure}
\centering
\includegraphics[width = 0.69\textwidth]{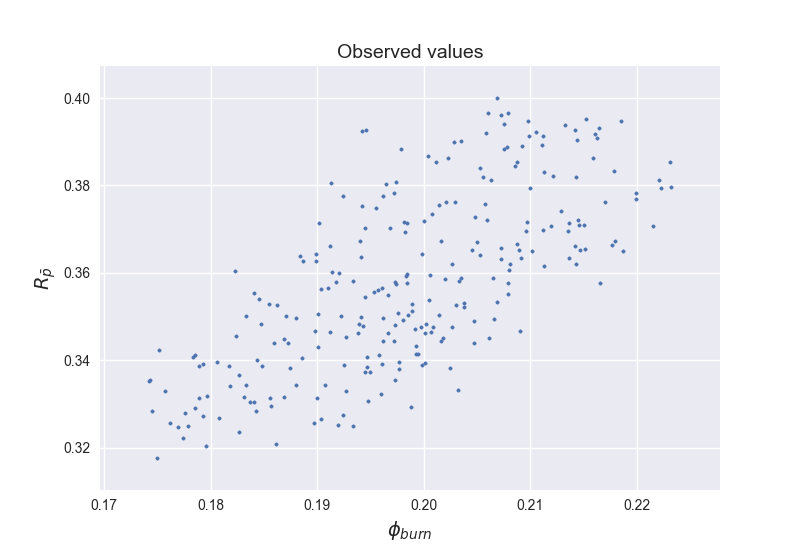}
\caption{Set of observations of the two quantities of interest. \label{fig:dataset}}
\end{figure}

PC expansions of $u_1 := \phi_{B}$ and $u_2 := R_{\bar{P}}$ of order
$Q = 4$ are constructed using Algorithm \ref{alg:CS_BA_2} for $d_0 =
1, 2, 3, 4$ and $5$ on a data set consisting of $256$ Monte Carlo
samples, shown in Fig. \ref{fig:dataset}. For each choice of $d_0$ and
for both QoIs, it is observed that the algorithm converges to a
solution after only 4-5 interchanges over the $\ell_1$-minimization
procedures. In addition, for each $d_0$ the cross validation procedure
is repeated independently in order to re-estimate $\epsilon$. As
$d_0$ increases, the set of values $\epsilon_{tr}^j$ is upper-bounded
by the value chosen at $d_0-1$, and so the value for $\epsilon$ decreases. This agrees with intuition which suggests that by increasing the dimensionality of the adapted expansion, we should expect the fit on data to improve.

\begin{figure}
\centering
\includegraphics[width = 0.49\textwidth]{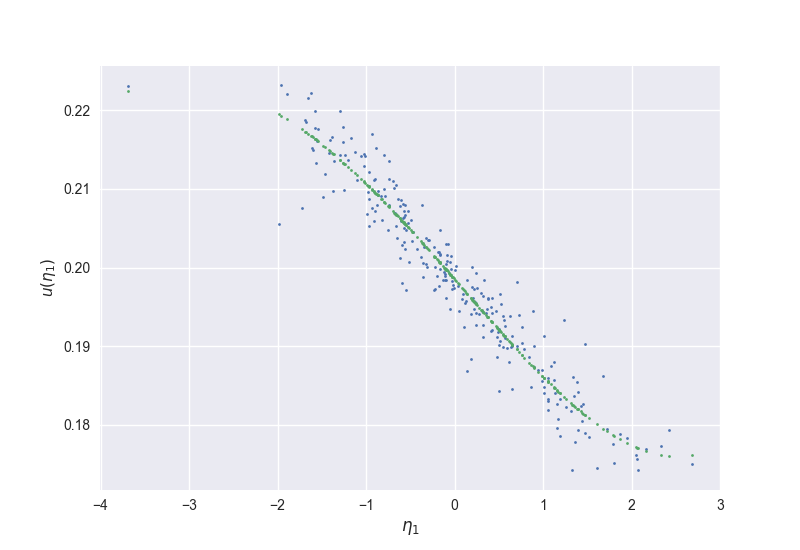}
\includegraphics[width = 0.49\textwidth]{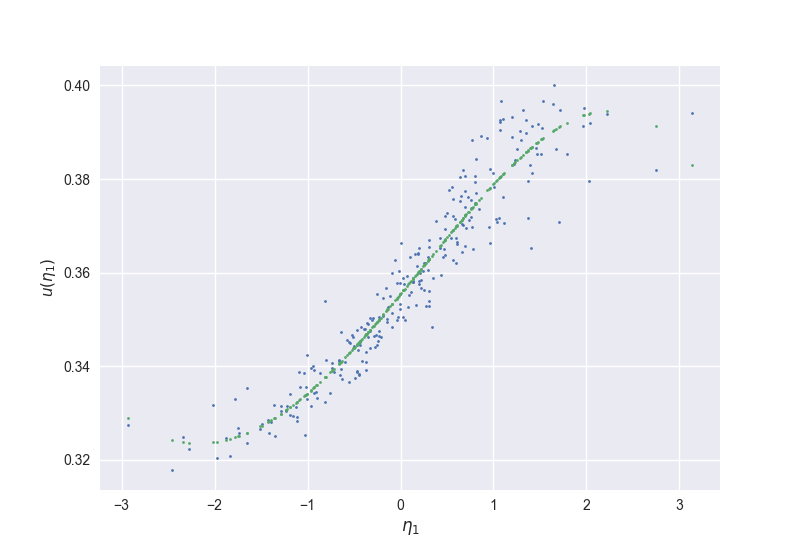}\\
\includegraphics[width = 0.49\textwidth]{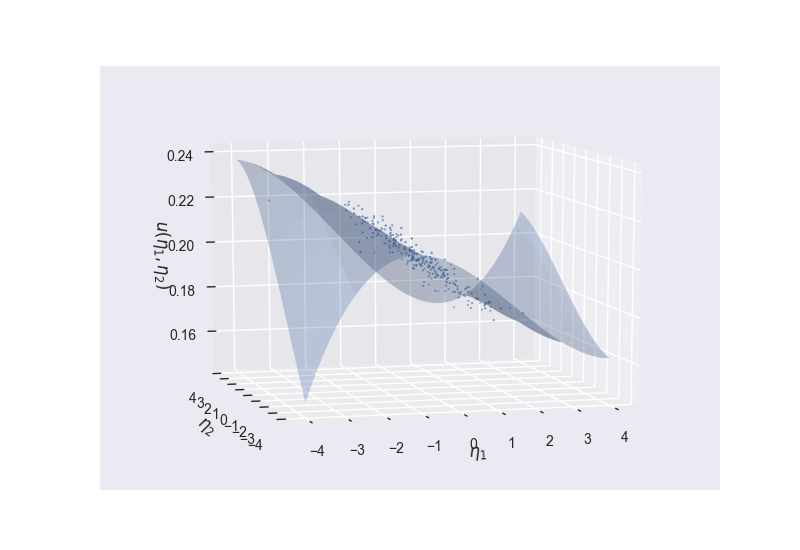}
\includegraphics[width = 0.49\textwidth]{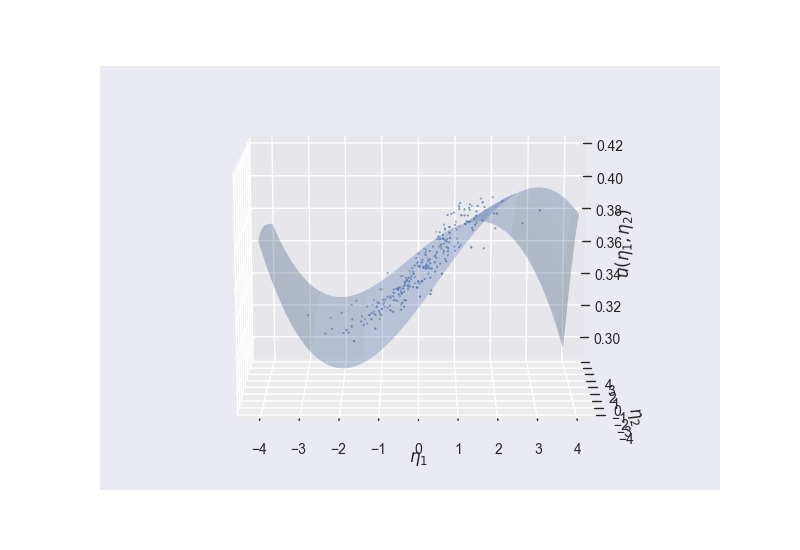}\\
\includegraphics[width = 0.49\textwidth]{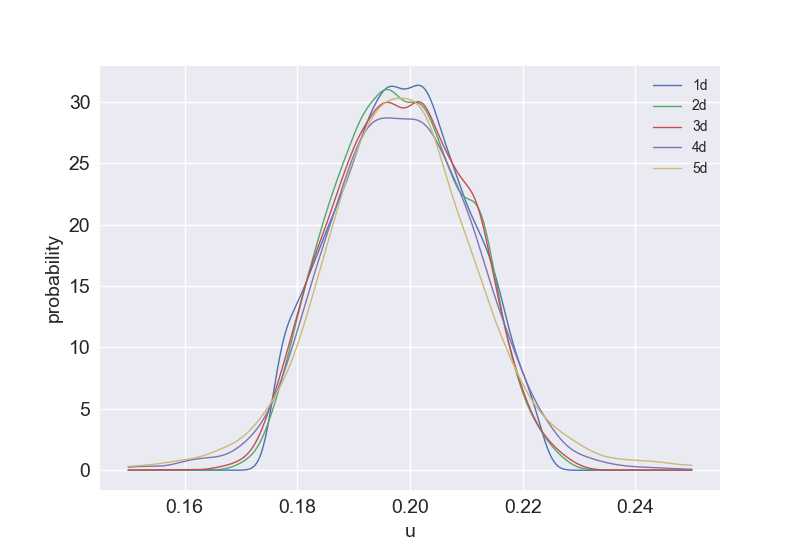}
\includegraphics[width = 0.49\textwidth]{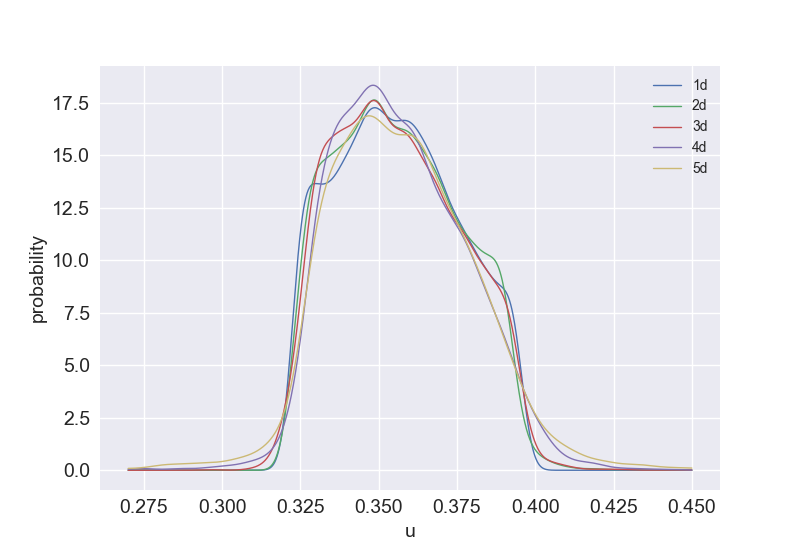}
\caption{Top: $1$d chaos expansions with rotated inputs evaluated at the input data points. Middle: $2$d chaos expansions with rotated inputs evaluated over the $[-4,4]^2$ domain. The data points in both cases are plotted as functions of the rotated inputs. Bottom: Density functions for up to $5$d chaos expansions. Left column results corresponds to $\phi_{B}$ while right column corresponds to $R_{\bar{P}}$. \label{fig:chaos_results}}
\end{figure}

\begin{figure}
\centering
\includegraphics[width = 0.49\textwidth]{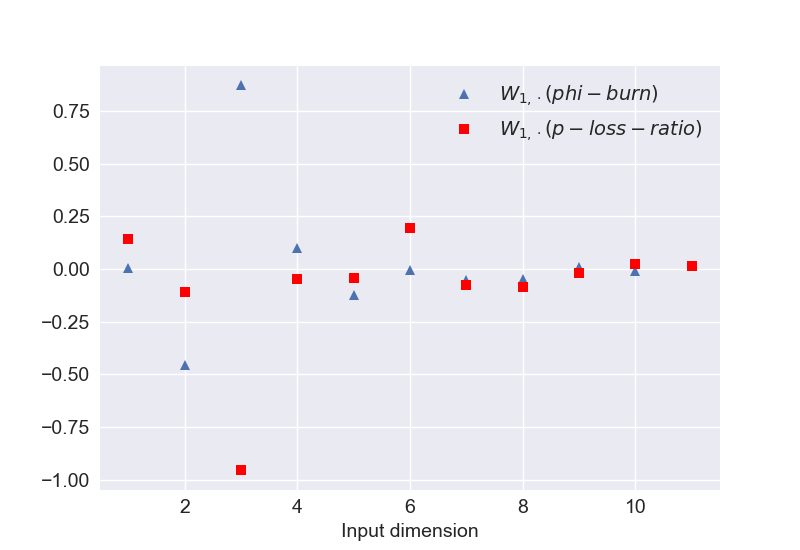}
\includegraphics[width = 0.49\textwidth]{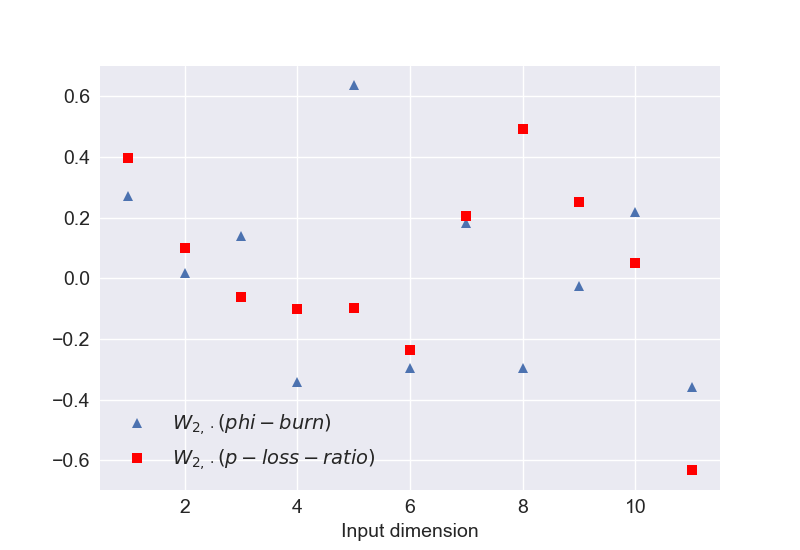}
\caption{Comparison of the values of the first (left) and second (right) rows of the projection matrix for the two QoIs. \label{fig:iso_results}}
\end{figure}

Fig. \ref{fig:chaos_results} shows plots of the resulting 1d and 2d adaptations for the QoIs along with density functions of the chaos
expansions up to 5d using all available observations. We observe that
the computed PC expansions provide a qualitatively good fit on the
observed data when the latter is plotted as function of the new
rotated, reduced basis. Hence, this indicates the QoIs under
consideration can be successfully represented as lower dimensional
functions of the new Gaussian germs, and capture the probabilistic
behavior of the full PC expansions. This is further supported by the
comparison of density functions of the $5$ PC expansions, which show
almost identical shape for both QoIs. Fig. \ref{fig:iso_results} shows
the values of the first two rows of computed projection matrices that
define $\eta_1$ and $\eta_2$ from $\bxi$, for each of the
QoIs. Assuming that a 1d or 2d expansion can be used as a functional
representation of each QoI, these values can be used as a measure of
sensitivity to each $\xi_i$ as each of the values determines the
impact of the corresponding $\xi_i$ on the variance of $\eta_1$ and
$\eta_2$. Overall we observe that the first row values weigh the
$\xi_i$'s in a similar way for the two QoIs. The values of the second
row are slightly different for each case, however several entries
maintain an agreement.

\begin{figure}
\centering
\includegraphics[width = 0.49\textwidth]{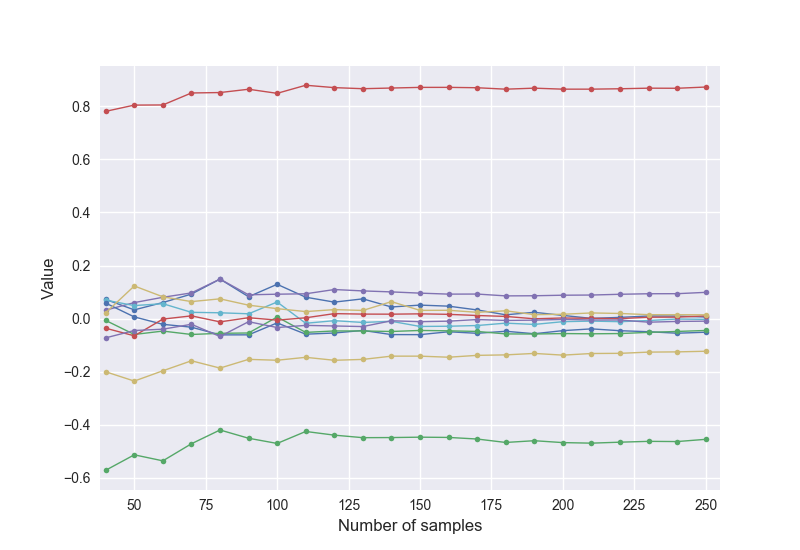}
\includegraphics[width = 0.49\textwidth]{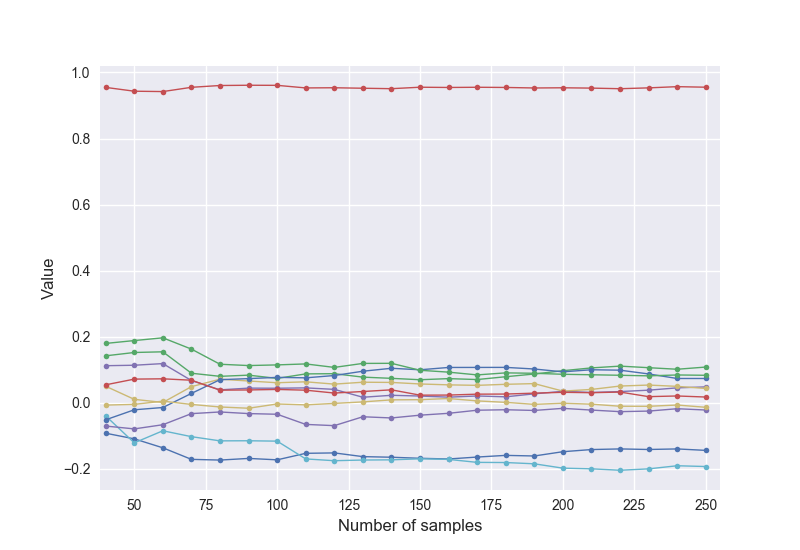}
\caption{Estimated projection vector values of the $1$d expansion versus number of samples used for $\phi_{B}$ (left) and $R_{\bar{P}}$ (right). \label{fig:iso_vs_samples}}
\end{figure}

\begin{figure}
\centering
\includegraphics[width = 0.49\textwidth]{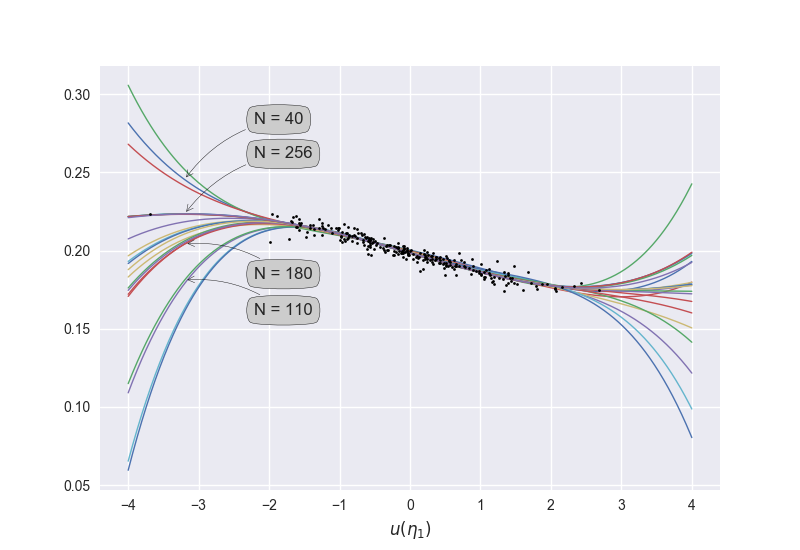}
\includegraphics[width = 0.49\textwidth]{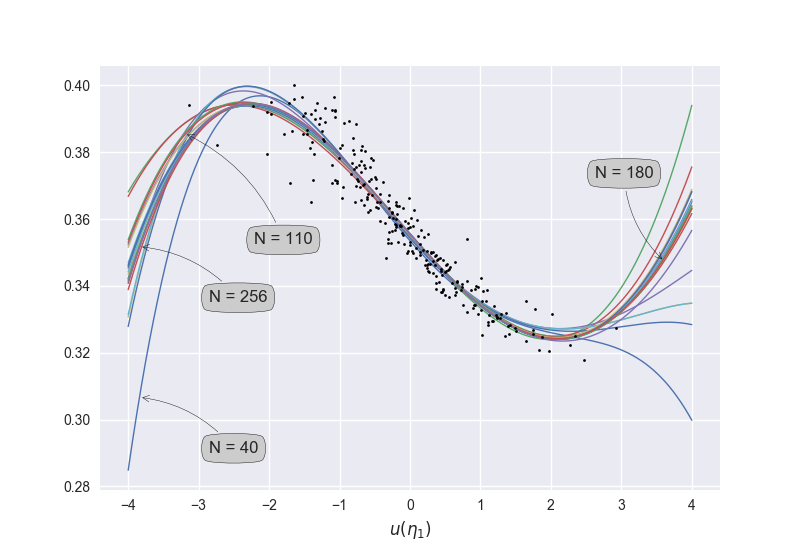}
\caption{Plots of all the $1$d PC expansions of $\phi_{B}$ (left) and $R_{\bar{P}}$ (right), computed for a number of samples varying from $N = 40$ to $N = 256$. All $256$ samples are also shown.  \label{fig:u_1d_all}}
\end{figure}

\begin{figure}
\centering
\includegraphics[width = 0.49\textwidth]{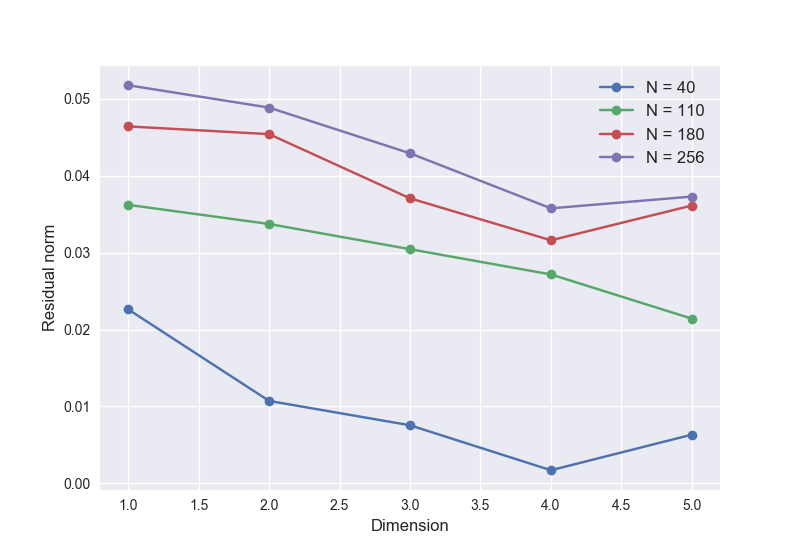}
\includegraphics[width = 0.49\textwidth]{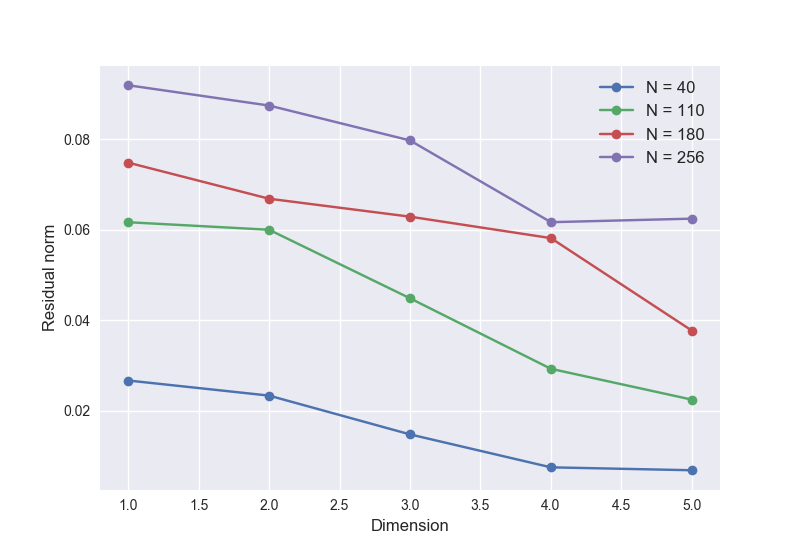}
\caption{Plots of $\ell_2$ residuals versus dimensionality with $N = 40, 110, 180$ and $256$ for $\phi_{B}$ (left) and $R_{\bar{P}}$ (right). \label{fig:l2_residual}}
\end{figure}

We also explore the dependence of the algorithm performance with
respect to the number of data samples. Fig. \ref{fig:iso_vs_samples}
shows the $11$ values of the projection vector $\bw$ for a 1d
adaptation when the number of samples varies from $40$ to $256$ with
$10$-sample batch being added at a time (and $16$ at the final step
from $240$ to $256$). One can observe small fluctuations in the values
when the samples vary from $40$ to over $100$ when they start to converge, which suggests that the isometry could be safely recovered with about $120$ samples. Next, all the computed $1$d expansions are shown in Fig. \ref{fig:u_1d_all}, along with all $256$ data points in order to assess the quality of the fit. Overall one can conclude that for both QoIs, even the expansions obtained with very few data points provide about the same fit on the data along the whole range of the Gaussian germ $\eta_1$ and they only start diverging from each other around the tails of the distribution that is past the $-2, 2$ values which correspond to areas where $\eta_1$ can be found with probability less than $0.05$. At last, plots of the $\ell_2$ errors $||\bu - \Psi_{\bW}\bc^*||^2_2$ versus the dimensionality of the chaos expansion are shown in Fig. \ref{fig:l2_residual} for $N = 40, 110, 180$ and $256$ verifying our initial intuition that the data fit should be improving as we move towards higher dimensions. 
Overall our approach has provided a thorough understanding and description of statistics of two complex and higly nonlinear quantities of interest in this computationally expensive study of turbulent combustion in the HiFIRE scramjet engine that would be otherwise infeasible, given this limited number of  available model evaluations. Accurate description of the probability distribution of the two QoI's had been achieved with as low as $120$ samples, whereas our 1-dimensional expansions and specifically the their projections indicate that the stagnation temperature and Mach number are the dominant parameters affecting $\phi_B$ and stagnation pressure and turbulence length scale are the two most dominant parameters affecting $R_{\bar{p}}$. Moreover, tasks such as computing joint distributions and joint probabilities of the QoI's can now be performed in an efficient manner due to the availability of their analytical representations with respect to a low dimensional input.

\section{Conclusions}
\label{sec:conclusions}

We presented a novel method for dimension reduction of polynomial
chaos expansions by combining compressive sensing with basis
adaptation. Starting with a low dimension, the new algorithm
finds an optimal rotated PC expansion by alternating between two
subproblems: computing the chaos coefficients via
$\ell_1$-minimization, and constructing an orthogonal rotation matrix
through $\ell_2$-minimization. The appropriate reduced dimension can
then be selected by assessing the convergence of data fit,
statistics, and distribution of the quantities of interest being
represented.

The main advantage of the new method is its efficiency in estimating
chaos expansions on a reduced dimension and with a usually
significantly smaller number of samples compared to a full dimensional
PC expansion.
It also advances the basis adaptation framework by coupling it with compressive sensing
algorithms, thus
offering flexibility to avoid the computational burden associated with
the use of quadrature methods for estimation of their coefficients in
pseudo-spectral approaches, particularly in high dimensions.

A promising future direction of this research would definitely involve
testing its applicability to more computationally expensive or
physically complex problems and to even higher dimensions. From a
theoretical point of view, our methodology can be shown to be a
special case of a Bayesian Compressive Sensing approach in the spirit
of the work of Sargsyan \cite{sargsyan_IJUQ} and our solution is the
\emph{maximum a posteriori} (MAP) estimate, corresponding to a Laplace
and Uniform (on the Stiefel manifold) priors on the coefficients and
the projection matrix respectively and a Gaussian Likelihood
function. More general prior and likelihood function choices could
lead to a more thorough understanding of the behavior of the MAP
solution and the computational challenges of the algorithm. Furthermore, techniques for sampling from distributions defined on Stiefel manifolds as in \cite{chowd} would enable estimation
of posterior distributions of projection matrices and would
be another step towards the ultimate goal that is the fully Bayesian
solution of the problem. Recent progress on this direction has shown that an approach for sampling from the marginal posteriors of the coefficients and the rotation matrix, using a combination of variational inference and hamiltonian Monte Carlo on the Stiefel manifold, can result in an efficient way of exploring the joint posterior \cite{tsilifis_geodesics}, though more generic approaches are yet to be developed.

\section{Acknowledgements}

Support for this research was provided by the Defense Advanced Research Projects Agency (DARPA) program on Enabling Quantification of Uncertainty in Physical Systems (EQUiPS).
This research used resources of the National Energy Research Scientific Computing Center (NERSC), a U.S. Department of Energy Office of Science User Facility operated under Contract No. DE-AC02-05CH11231.
Sandia National Laboratories is a multimission laboratory managed and
operated by National Technology and Engineering Solutions of Sandia,
LLC., a wholly owned subsidiary of Honeywell International, Inc., for
the U.S. Department of Energy's National Nuclear Security
Administration under contract DE-NA-0003525. 
The views expressed in the article do not necessarily represent the views of the U.S. Department Of Energy or the United States Government.

\appendix 

\section{Gradient of the $\ell_2$ error}
\label{sec:log_like_grad}

Here we seek to derive the gradient of the $\ell_2$ error function
\begin{equation}
\bJ(\bW) = ||\bu - \Psi_{\bW} \bc||_2^2 = \left(\bu - \Psi_{\bW}\bc\right)^T\left(\bu - \Psi_{\bW}\bc\right)
\end{equation}
with respect to the entries of $\bW$. For arbitraty $\theta := (\bW)_{ij}$ we get
\begin{equation}
\frac{\partial \bJ(\bW)}{\partial \theta} = - 2\bu^T \frac{\partial \Psi_{\bW} }{\partial \theta}\bc + \bc^T\left(\Psi_{\bW}^T \frac{\partial \Psi_{\bW}}{\partial \theta} + \frac{\partial \Psi_{\bW}}{\partial \theta}^T \Psi_{\bW} \right) \bc.
\end{equation}
where $\frac{\partial \Psi_{\bW}}{\partial \theta}$ is the matrix with entries $\frac{\partial(\Psi_{\bW})_{k\bbeta}}{\partial \theta} = \frac{\partial \psi_{\bbeta}(\bW \bxi^{(k)})}{\partial \theta}$. For each multi-index $\bbeta$ and for $\epsilon_i$ being the multi-index with value $1$ as its $i$-th entry and zero elsewhere, we have \begin{equation}
\frac{\partial \psi_{\bbeta}(\bW \bxi)}{\partial \theta} = \sum_{r = 1}^{d_0} \frac{\partial \psi_{\bbeta}}{\partial \eta_r}\frac{\partial \eta_r}{\partial \theta} = \frac{\partial \psi_{\bbeta}}{\partial \eta_i}\frac{\partial \eta_i}{\partial \theta} = \sqrt{\beta_i}\psi_{\bbeta-\epsilon_i}(\bfeta) \xi_j
\end{equation}
where the last equality makes use of the fact that $\psi_n'(\eta) = \sqrt{n}\psi_{n-1}(\eta)$ and that gives
\begin{equation}
\frac{\partial}{\partial \eta_i}\psi_{\bbeta}(\bfeta) = \psi_{\beta_i}'(\eta_i)\prod_{\substack{j=1 \\ j\neq i}} \psi_{\beta_j}(\eta_j) = \sqrt{\beta_i} \psi_{\bbeta - \epsilon_i}(\bfeta).
\end{equation}

\section{Bayesian problem formulation and \emph{maximum a posteriori} solutions}
\label{sec:map}

In order to clarify the motivation for our proposed methodology that was used to estimate the coefficients $\bc$ and the projection matrix $\bW$, we reformulate the problem in a Bayesian setting. As our stating point, we treat both $\bc$ and $\bW$ as random quantities that are assigned \emph{prior} distributions, say $p(\bc, \bW) = p(\bc)p(\bW)$, where clearly the factorization implies independence between $\bc$ and $\bW$. The prior can then be updated to a posterior distribution conditioned on available data $\calD$ given by
\begin{equation}
p(\bc, \bW | \calD) \propto \calL(\calD | \bc, \bW) p(\bc) p(\bW),
\end{equation}
where the data is defined as $\calD = \left\{\{\bxi^{(i)}\}_{i=1}^N , \{\hat{u}^{(i)}\}_{i=1}^N\right\}$, that is the set of available model input/output pairs. The likelihood considered here is Gaussian 
\begin{equation}
\calL(\calD | \bc, \bW) = \left(2\pi \sigma^2\right)^{-N/2} \exp\left\{ - \frac{1}{2\sigma^2} ||\bu - \Psi_{\bW} \bc||_2^2 \right\},
\end{equation}
where $\bu = (\hat{u}^{(1)}, \dots, \hat{u}^{(N)})^T$ is the vector of
output data, $\{\bxi^{(i)}\}_{i=1}^N$ is the set of input points
corresponding to the data outputs, and $\Psi$ is the measurement
matrix with entries $(\Psi_{\bW})_{ij} = \psi_{j}(\bW\bxi^{(i)})$, $i = 1,\dots, N$, $j\in \calJ_Q^d$. 

For the coefficients we assume a Laplace prior that has been commonly used \cite{ji, sargsyan_IJUQ} as a sparsity-inducing prior and is given as 
\begin{equation}
p(\bc) = (\tau / 2)^{\vert \calJ_Q \vert+ 1} \exp\left\{ - \tau \sum_{i = 0}^{\vert \calJ_Q\vert} |c_i|\right\},
\end{equation}
while for $\bW$, provided that naturally one has no prior information regarding the optimal projection, we choose a uniform prior. Taking into account the pairwise orthogonality constraints among the rows of $\bW$, the probability measure is defined on the Stiefel manifold $\calM_{d_0}^d$ and its constant density is 
\begin{equation}
p(\bW) = \frac{\Gamma_{d_0}(d / 2)}{2^{d_0}\pi^{d_0 d/2}},
\end{equation}
where $\Gamma_{d_0}(\cdot)$ is the $d_0$-variate Gamma function \cite{chikuse}. Combining the above priors and the likelihood together, one can easily see that the \emph{maximum a posteriori (MAP)} estimate satisfies 
\begin{eqnarray}
\bc^*, \bW^* & = & \arg\max_{\bc, \bW}\{- \calF(\bc, \bW)\} = \arg\min_{\bc, \bW} \{\calF(\bc, \bW) \}
\end{eqnarray}
where 
\begin{equation}
\label{eq:minim}
\calF(\bc, \bW) \propto - \log \calL(\calD | \bc, \bW) + \tau ||\bc||_1  = \frac{1}{2\sigma^2}||\bu - \Psi_{\bW} \bc||_2^2 + \tau ||\bc||_1,
\end{equation}

Note that for the case where standard PC expansions are used and no projection matrix is involved, optimization of the objective function $\calF$ above, with respect to $\bc$ only, reduces to the classical compressive sensing problem or its Bayesian counterpart \cite{ji, sargsyan_IJUQ}. In the general case however, one way of making the optimization problem tractable, is to employ a coordinate descent algorithm \cite{wright2015} that breaks the problem into two easier subproblems that are solved within a loop as described in Algorithm \ref{alg:coord}.

\begin{algorithm}[h]
\caption{Coordinate descent for minimization of (\ref{eq:minim}) \label{alg:coord}}
Choose $\bc_0$, $\bW_0$, tolerance $\textrm{tol}$, set $n = 0$\\
\While{Change in $\calF$ is less than \textrm{tol}}{
$\bc_{n+1} \longleftarrow \arg\min_{\bc}\left\{\calF(\bc_n, \bW_n)\right\}$\\
$\bW_{n+1} \longleftarrow \arg\min_{\bW}\left\{\calF(\bc_{n+1}, \bW_n)\right\}$\\
$n \leftarrow n+1$
 }
Return $\bc^*, \bW^* := \bc_n, \bW_n$. 
\end{algorithm}

As mentioned above, even when $\bW$ is present in $\Psi_{\bW}$, minimization of $\calF(\bc,\bW)$ with respect to $\bc$ only, is the lagrangian version of the LASSO problem, or equivalently, the $\ell_1$-minimization problem. On the other hand, the $\calF(\bc,\bW)$ is minimized with respect to $\bW$, the regularization term $\tau ||\bc||_1$ can be ignored and the problem reduces to minimizing the misfit term $\big\Vert\bu - \Psi_{\bW}\bc\big\Vert_2^2$ or equivalently, the negative log-likelihood. In other words, Algorithm \ref{alg:coord} becomes Algorithm \ref{alg:CS_BA} and that justifies the heuristic two-step optimization problem in our methodology.

\section{On the equivalence of solutions of the $\ell_2$ error}
\label{sec:equiv}

Here we explain in some more detail the connection between Algorithms \ref{alg:CS_BA} and \ref{alg:CS_BA_2}. Assume that $\bc^*$ and $\bW^*$ are the outcome of Alg. \ref{alg:CS_BA}. For the corresponding expansion
\begin{equation}
u(\tilde{\bfeta}) = \sum_{\balpha} c_{\alpha} \psi_{\balpha}(\tilde{\bfeta})
\end{equation}
with $\tilde{\bfeta} = \bW^*\bxi$, let $\bB$ be any $d_0\times d_0$ isometry matrix and set $\bzeta = \bB \tilde{\bfeta}$. The expansion can be rewritten as
\begin{equation}
u(\bfeta) \stackrel{a.s.}{=} \hat{u}(\bzeta) = \sum_{\bbeta} \hat{c}_{\bbeta} \psi_{\bbeta}(\bzeta).
\end{equation}
Denoting with $\hat{\bc}$ the vector of new coefficients and with $\Psi_{\bB\bW^*}$ the new measurement matrix and using the almost sure equality of the two expansions we get that
\begin{equation}
\min_{\bW}||\bu - \Psi_{\bW}\bc||_2 = ||\bu - \Psi_{\bW^*}\bc^*||_2 = ||\bu - \Psi_{\bB\bW^*}\hat{\bc}||_2
\end{equation}
that is $\bB\bW^*$ provides the same fit on the data. Intuitively, Alg. \ref{alg:CS_BA} can attain a particular minimum for many different rotation matrices, depending each time on the starting values of $\bW$ which in general are chosen randomly. Equality of the corresponding coefficients of course is not quaranteed. Each set of coefficients attains a minimum $\ell_1$ norm only for the corresponding $\Psi_{\bW}$.

\section*{}
\bibliographystyle{plain}
\bibliography{references}

\begin{thebibliography}{10}

\bibitem{Doostan:2007}
Doostan A., Ghanem R., and J.~Red-Horse.
\newblock Stochastic model reduction for chaos representations.
\newblock {\em Computer Methods in Applied Mechanics and Engineering},
  196(37-40):3951--3966, 2007.

\bibitem{Arnst:2012}
M.~Arnst, R.~Ghanem, E.~Phipps, , and J.~Red-Horse.
\newblock Measure transformation and efficient quadrature in
  reduced-dimensional stochastic modeling of coupled problems.
\newblock {\em International Journal for Numerical Methods in Engineering},
  92(12):1044--1080, 2012.

\bibitem{babuska}
I.~Babu\u{s}ka, F.~Nobile, and R.~Tempone.
\newblock A stochastic collocation method for elliptic partial differential
  equations with random input data.
\newblock {\em SIAM Journal on Numerical Analysis}, 45:1005--1034, 2007.

\bibitem{bellman}
R.E. Bellman.
\newblock {\em Adaptive control processes: a guided tour (Vol. 2045)}.
\newblock Princeton university press, 2015.

\bibitem{blatman}
G.~Blatman and B.~Sudret.
\newblock Adaptive sparse polynomial chaos expansion based on least angle
  regression.
\newblock {\em Journal of Computational Physics}, 230:2345--2367, 2011.

\bibitem{cameron}
R.~Cameron and W.~Martin.
\newblock The orthogonal development of nonlinear functionals in series of
  fourier-hermite functionals.
\newblock {\em Annals of Mathematics}, 48:385--392, 1947.

\bibitem{Candes2006a}
Emmanuel~J. Cand{\`{e}}s, Justin Romberg, and Terence Tao.
\newblock Robust uncertainty principles: Exact signal reconstruction from
  highly incomplete frequency information.
\newblock {\em IEEE Transactions on Information Theory}, 52(2):489--509, 2006.

\bibitem{chikuse}
Y.~Chikuse.
\newblock {\em Statistics on special manifolds (Vol. 174)}.
\newblock Springer Science \& Business Media, 2012.

\bibitem{chowd}
K.~Chowdhary and H.N. Najm.
\newblock Bayesian estimation of {K}arhunen-€"{L}o\`{e}ve expansions; a random
  subspace approach.
\newblock {\em Journal of Computational Physics}, 319:280--293, 2016.

\bibitem{DR_algo_2}
P.L. Combettes and J.C. Pesquet.
\newblock A {D}ouglas - {R}achford splitting approach to nonsmooth convex
  variational signal recovery.
\newblock {\em IEEE Journal of Selected Topics in Signal Processing},
  1:564--574, 2007.

\bibitem{constantine}
P.~G. Constantine, E.~Dow, and Q.~Wang.
\newblock Active {S}ubspace methods in theory and practice: {A}pplications to
  kriging surfaces.
\newblock {\em SIAM Journal on Scientific Computing}, 36:A1500--A1524, 2014.

\bibitem{constantine_hyshot}
P.G. Constantine, M.~Emory, J.~Larsson, and G.~Iaccarino.
\newblock Exploiting active subspaces to quantify uncertainty in the numerical
  simulation of the hyshot ii scramjet.
\newblock {\em Journal of Computational Physics}, 302:1--20, 2015.

\bibitem{Dolvin2008}
D.~Dolvin.
\newblock Hypersonic {I}nternational {F}light {R}esearch and {E}xperimentation
  ({H}{I}{F}i{R}{E}) fundamental science and technology development strategy.
\newblock In {\em 15th AIAA International Space Planes and Hypersonic Systems
  and Technologies Conference (p. 2581)}, Dayton, OH, April 2008.

\bibitem{Dolvin2009}
D.~Dolvin.
\newblock Hypersonic {I}nternational {F}light {R}esearch and {E}xperimentation
  {T}echnology {D}evelopment and {F}light {C}ertification {S}trategy.
\newblock In {\em 16th AIAA/DLR/DGLR International Space Planes and Hypersonic
  Systems and Technologies Conference (p. 7228)}, Bremen, Germany, 2009.

\bibitem{donoho}
D.L. Donoho.
\newblock Compressed sensing.
\newblock {\em IEEE Transactions on information theory}, 52:1289--1306, 2006.

\bibitem{donoho_elad}
D.L. Donoho, M.~Elad, and V.N. Temlyakov.
\newblock Stable recovery of sparse overcomplete representations in the
  presence of noise.
\newblock {\em IEEE Transactions on information theory}, 52:6--18, 2006.

\bibitem{doostan_icc}
A.~Doostan and G.~Iaccarino.
\newblock A least-squares approximation of partial differential equations with
  high-dimensional random inputs.
\newblock {\em Journal of Computational Physics}, 228:4332--4345, 2009.

\bibitem{doostan_owhadi}
A.~Doostan and H.~Owhadi.
\newblock A non-adapted sparse approximation of pdes with stochastic inputs.
\newblock {\em Journal of Computational Physics}, 230:3015--3034, 2011.

\bibitem{DR_algo_1}
J.~Douglas and H.H. Rachford.
\newblock On the numerical solution of heat conduction problems in two and
  three space variables.
\newblock {\em Transactions of the American mathematical Society}, 82:421--439,
  1956.

\bibitem{ghanem_wrr}
R.~Ghanem.
\newblock Scales of fluctuation and the propagation of uncertainty in random
  porous media.
\newblock {\em Water Resources Research}, 34:2123--2136, 1998.

\bibitem{ghanem}
R.~Ghanem.
\newblock Ingredients for a general purpose stochastic finite elements
  implementation.
\newblock {\em Computer Methods in Applied Mechanics and Engineering},
  168:19--34, 1999.

\bibitem{ghanem_dham}
R.~Ghanem and S.~Dham.
\newblock Stochastic finite element analysis for multiphase flow in
  heterogeneous porous media.
\newblock {\em Transport in Porous Media}, 32:239--262, 1998.

\bibitem{ghanem_redhorse}
R.~Ghanem and J.~Red-Horse.
\newblock Propagation of probabilistic uncertainty in complex physical systems
  using a stochastic finite element approach.
\newblock {\em Physica D: Nonlinear Phenomena}, 133:137--144, 1999.

\bibitem{ghanem_spanos}
R.~Ghanem and P.~Spanos.
\newblock {\em Stochastic finite elements: A spectral approach}.
\newblock Springer-Verlag, 1991.

\bibitem{Ghosh:2008}
D.~Ghosh and R.~Ghanem.
\newblock Stochastic convergence acceleration through basis enrichment of
  polynomial chaos expansions.
\newblock {\em International Journal of Numerical methods in Engineering},
  73(2):162--184, 2008.

\bibitem{Gruber2008}
Mark~R. Gruber, Kevin Jackson, and Jiwen Liu.
\newblock {Hydrocarbon-Fueled Scramjet Combustor Flowpath Development for Mach
  6-8 HIFiRE Flight Experiments}.
\newblock Technical report, AFRL, 2008.

\bibitem{hass}
N.~Hass, K.~Cabell, A.~Storch, and M.~Gruber.
\newblock {H}{I}{F}i{R}{E} direct-connect rig ({H}{D}{C}{R}) phase {I} scramjet
  test results from the {N}{A}{S}{A} {L}angley arc-heated {S}cramjet test
  facility.
\newblock In {\em 17th AIAA international space planes and hypersonic systems
  and technologies conference (p. 2248)}, 2011.

\bibitem{hokanson}
J.M. Hokanson and P.G. Constantine.
\newblock Data-driven polynomial ridge approximation using variable projection.
\newblock {\em SIAM Journal on Scientific Computing}, 40:A1566--A1589, 2018.

\bibitem{huan_sobol}
X.~Huan, C.~Safta, K.~Sargsyan, G.~Geraci, M.S. Eldred, Z.P. Vane, G.~Lacaze,
  J.C. Oefelein, and H.N. Najm.
\newblock Global {S}ensitivity {A}nalysis and {E}stimation of {M}odel {E}rror,
  toward {U}ncertainty {Q}uantification in {S}cramjet {C}omputations.
\newblock {\em AIAA Journal}, 56:1170--1184, 2018.

\bibitem{huan_CS}
X.~Huan, C.~Safta, K.~Sargsyan, Z.P. Vane, G.~Lacaze, J.C. Oefelein, and H.N.
  Najm.
\newblock Compressive {S}ensing with {C}ross-{V}alidation and {S}top-{S}ampling
  for {S}parse {P}olynomial {C}haos {E}xpansions.
\newblock {\em SIAM/ASA Journal on Uncertainty Quantification}, 6:907--936,
  2018.

\bibitem{jackson}
K.R. Jackson, M.R. Gruber, and S.~Buccellato.
\newblock H{I}{F}i{R}{E} {F}light 2 {O}verview and {S}tatus {U}pdate 2011.
\newblock In {\em 17th AIAA International Space Planes and Hypersonic Systems
  and Technologies Conference}, pages 2011--2202, San Francisco, CA, 2011.

\bibitem{jakeman}
J.D. Jakeman, M.S. Eldred, and K.~Sargsyan.
\newblock Enhancing $\ell_1$-minimization estimates of polynomial chaos
  expansions using basis selection.
\newblock {\em Journal of Computational Physics}, 289:18--34, 2015.

\bibitem{ji}
S.~Ji, Y.~Xue, and L.~Carin.
\newblock Bayesian compressive sensing.
\newblock {\em IEEE Transactions on Signal Processing}, 56:2346, 2008.

\bibitem{Lacaze2017}
G.~Lacaze, Z.~Vane, and J.C. Oefelein.
\newblock Large {E}ddy {S}imulation of the {H}{I}{F}i{R}{E} {D}irect {C}onnect
  {R}ig {S}cramjet {C}ombustor.
\newblock In {\em 55th AIAA Aerospace Sciences Meeting (p. 0142)}, Grapevine,
  TX, 2017.

\bibitem{le_maitre_etal}
O.P. Le~Ma\^{i}tre, M.T. Reagan, H.N. Najm, R.G. Ghanem, and O.M. Knio.
\newblock A stochastic projection method for fluid flow: Ii. random process.
\newblock {\em Journal of Computational Physics}, 181:9--44, 2002.

\bibitem{marzouk_etal}
Y.~M. Marzouk, H.~N. Najm, and L.~Rahn.
\newblock Stochastic spectral methods for efficient bayesian solution of
  inverse problems.
\newblock {\em Journal of Computational Physics}, 224:560--586, 2007.

\bibitem{najm}
H.N. Najm.
\newblock Uncertainty quantification and polynomial chaos techniques in
  computational fluid dynamics.
\newblock {\em Annual Review of Fluid Mechanics}, 41:35--52, 2009.

\bibitem{nocedal}
J.~Nocedal and S.~Wright.
\newblock {\em Numerical optimization}.
\newblock Springer Science \& Business Media, 2006.

\bibitem{oefelein_phd}
J.C. Oefelein.
\newblock {\em Simulation and analysis of turbulent multiphase combustion
  processes at high pressures}.
\newblock PhD thesis, Pennsylvania State University, University Park,
  Pennsylvania, May 1997.

\bibitem{oefelein}
J.C. Oefelein.
\newblock Large eddy simulation of turbulent combustion processes in propulsion
  and power systems.
\newblock {\em Progress in Aerospace Sciences}, 42:2--37, 2006.

\bibitem{pearson}
K.~Pearson.
\newblock Liii. on lines and planes of closest fit to systems of points in
  space.
\newblock {\em The London, Edinburgh, and Dublin Philosophical Magazine and
  Journal of Science}, 2:559--572, 1901.

\bibitem{Pellett2007a}
G.~Pellett, S.~Vaden, and L.~Wilson.
\newblock Opposed jet burner extinction limits: simple mixed hydrocarbon
  scramjet fuels vs air.
\newblock In {\em 43rd AIAA/ASME/SAE/ASEE Joint Propulsion Conference \&
  Exhibit (p. 5664)}, July 2007.

\bibitem{peng_l1}
J.~Peng, J.~Hampton, and A.~Doostan.
\newblock A weighted ℓ1-minimization approach for sparse polynomial chaos
  expansions.
\newblock {\em Journal of Computational Physics}, 267:92--111, 2014.

\bibitem{rasmussen}
C.E. Rasmussen and C.K. Williams.
\newblock {\em Gaussian processes for machine learning (Vol. 1)}.
\newblock Cambridge: MIT press, 2006.

\bibitem{reagan}
M.T. Reagan, H.N. Najm, R.G. Ghanem, and O.M. Knio.
\newblock Uncertainty quantification in reacting-flow simulations through
  non-intrusive spectral projection.
\newblock {\em Combustion and Flame}, 132:545--555, 2003.

\bibitem{saltelli}
A.~Saltelli, M.~Ratto, T.~Andres, F.~Campolongo, J.~Cariboni, D.~Gatelli,
  M.~Saisana, and S.~Tarantola.
\newblock {\em Global sensitivity analysis: the primer}.
\newblock John Wiley \& Sons, 2008.

\bibitem{sargsyan_IJUQ}
K.~Sargsyan, C.~Safta, H.N. Najm, B.J. Debusschere, D.~Ricciuto, and
  P.~Thornton.
\newblock Dimensionality reduction for complex models via bayesian compressive
  sensing.
\newblock {\em International Journal for Uncertainty Quantification}, 4, 2014.

\bibitem{spalart}
P.R. Spalart, W.H. Jou, M.~Strelets, and S.R. Allmaras.
\newblock Comments on the feasibility of {LES} for wings, and on a hybrid
  {RANS/LES} approach.
\newblock {\em Advances in {DNS/LES}}, 1:4--8, 1997.

\bibitem{Storch2011}
A.~Storch, M.~Bynum, J.~Liu, and M.~Gruber.
\newblock Combustor operability and performance verification for
  {H}{I}{F}i{R}{E} flight 2.
\newblock In {\em 17th AIAA International Space Planes and Hypersonic Systems
  and Technologies Conference (p. 2249)}, San Francisco, CA, April 2011.

\bibitem{thimmisetty}
C.~Thimmisetty, P.~Tsilifis, and R.~Ghanem.
\newblock Homogeneous chaos basis adaptation for design optimization under
  uncertainty: Application to the oil well placement problem.
\newblock {\em AI EDAM}, 31:265--276, 2017.

\bibitem{tibshirani}
R.~Tibshirani.
\newblock Regression shrinkage and selection via the lasso.
\newblock {\em Journal of the Royal Statistical Society. Series B
  (Methodological)}, pages 267--288, 1996.

\bibitem{tipireddy}
R.~Tipireddy and R.G. Ghanem.
\newblock Basis adaptation in homogeneous chaos spaces.
\newblock {\em Journal of Computational Physics}, 259:304--317, 2014.

\bibitem{tripathy}
R.~Tripathy, I.~Bilionis, and M.~Gonzalez.
\newblock Gaussian processes with built-in dimensionality reduction:
  Applications to high-dimensional uncertainty propagation.
\newblock {\em Journal of Computational Physics}, 321:191--223, 2016.

\bibitem{tsilifis}
P.~Tsilifis and R.G. Ghanem.
\newblock Reduced wiener chaos representation of random fields via basis
  adaptation and projection.
\newblock {\em Journal of Computational Physics}, 341:102--120, 2017.

\bibitem{tsilifis_geodesics}
P.~Tsilifis and R.G. Ghanem.
\newblock Bayesian adaptation of chaos representations using variational
  inference and sampling on geodesics.
\newblock {\em Proc. R. Soc. A}, 474:20180285, 2018.

\bibitem{tsilifis_design}
P.~Tsilifis, R.G. Ghanem, and P.~Hajali.
\newblock Efficient bayesian experimentation using an expected information gain
  lower bound.
\newblock {\em SIAM/ASA Journal on Uncertainty Quantification}, 5:30--62, 2017.

\bibitem{tsilifis_gradient}
P.A. Tsilifis.
\newblock Gradient-informed basis adaptation for legendre chaos expansions.
\newblock {\em Journal of Verification, Validation and Uncertainty
  Quantification}, 3:011005, 2018.

\bibitem{wen}
Zaiwen Wen and Wotao Yin.
\newblock A feasible method for optimization with orthogonality constraints.
\newblock {\em Mathematical Programming}, 142:397--434, 2013.

\bibitem{wiener}
N.~Wiener.
\newblock The homogeneous chaos.
\newblock {\em American Journal of Mathematics}, 60:897--936, 1938.

\bibitem{Witteveen2011}
J.~Witteveen, K.~Duraisamy, and G.~Iaccarino.
\newblock Uncertainty {Q}uantification and error estimation in {S}cramjet
  simulation.
\newblock In {\em 17th AIAA International Space Planes and Hypersonic Systems
  and Technologies Conference (p. 2283)}, April 2011.

\bibitem{wright2015}
Stephen~J. Wright.
\newblock Coordinate descent algorithms.
\newblock {\em Mathematical Programming}, 151(1):3--34, Jun 2015.

\bibitem{xiu_fluid}
D.~Xiu and G.E. Karniadakis.
\newblock Modeling uncertainty in flow simulations via generalized polynomial
  chaos.
\newblock {\em Journal of Computational Physics}, 187:137--167, 2003.

\bibitem{yang_sparsity}
X.~Yang, H.~Lei, N.A. Baker, and G.~Lin.
\newblock Enhancing sparsity of hermite polynomial expansions by iterative
  rotations.
\newblock {\em Journal of Computational Physics}, 307:94--109, 2016.

\bibitem{yentsch}
R.J. Yentsch and D.V. Gaitonde.
\newblock Exploratory simulations of the hifire 2 scramjet flowpath.
\newblock In {\em 48th AIAA/ASME/SAE/ASEE Joint Propulsion Conference and
  Exhibit}, July 2012.

\end{thebibliography}

\end{document}